\PassOptionsToPackage{table,xcdraw}{xcolor}

\documentclass[10pt,twocolumn,letterpaper]{article}

\usepackage{cvpr}              

\usepackage{booktabs}
\usepackage{graphicx}
\usepackage{multirow}
\usepackage{makecell}
\usepackage{amssymb}
\usepackage{amsmath}
\usepackage[linesnumbered,ruled,vlined]{algorithm2e}
\usepackage{pifont}

\usepackage{listings}

\newcommand{\name}[0]{OmniTrack}

\makeatletter
\newcommand{\algorithmfootnote}[2][\footnotesize]{%
  \let\old@algocf@finish\@algocf@finish
  \def\@algocf@finish{\old@algocf@finish
    \leavevmode\rlap{\begin{minipage}{\linewidth}
    #1#2
    \end{minipage}}%
  }%
}
\makeatother

\makeatletter
\newcommand{\shortcline}[1]{%
  \noalign{\global\advance\@totalleftmargin by 0.15em}
  \noalign{\global\advance\linewidth by -10em}
  \cline{#1}%
  \noalign{\global\advance\@totalleftmargin by -0.5em}
  \noalign{\global\advance\linewidth by 1em}
}
\makeatother

\definecolor{codegreen}{rgb}{0.0,0.6,0.0}
\definecolor{mygray}{gray}{.9}
\definecolor{mygray1}{gray}{.7}
\definecolor{tabgray}{rgb}{0.957,0.945,0.925}

\newcommand{\car}{\includegraphics[width=4mm]{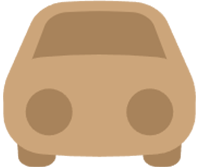}}
\newcommand{\robot}{\includegraphics[width=4mm]{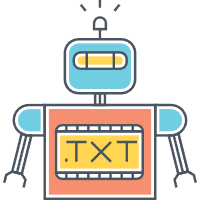}}
\newcommand{\robotdog}{\includegraphics[width=4mm]{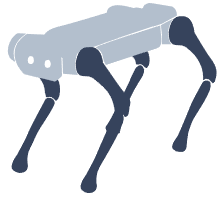}}
\newcommand{\webm}{\includegraphics[width=3.6mm]{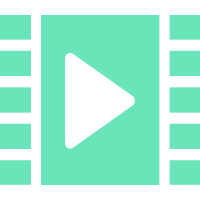}}
\newcommand{\mywebm}{\includegraphics[width=3.5mm]{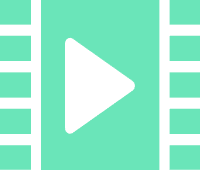}}

\newcommand{\wheels}{\includegraphics[width=4mm]{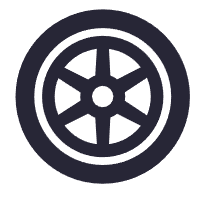}}
\newcommand{\gait}{\includegraphics[width=4mm]{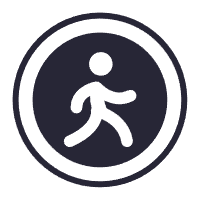}}
\newcommand{\stationary}{\includegraphics[width=4mm]{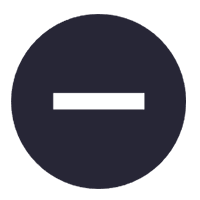}}

\newcommand{\mycheckmark}{\includegraphics[width=3.5mm]{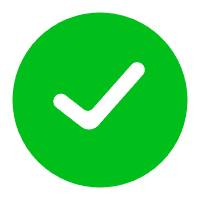}}
\newcommand{\crossmark}{\includegraphics[width=3.5mm]{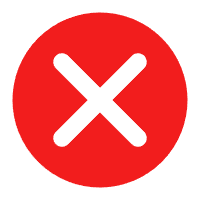}}

\newcommand{\topline}{\noalign{\hrule height 0.8 pt}} 
 
\newcommand{\bottomline}{\noalign{\hrule height 0.8 pt}} 

\definecolor{codebackgroundcolor}{RGB}{248,242,222}

\definecolor{r1}{rgb}{0.7569,0.2275,0.1294}
\definecolor{r2}{rgb}{0.0,0.5451,0.5451}
\definecolor{r3}{rgb}{0.5451,0.0,0.5451}

\definecolor{cvprblue}{rgb}{0.21,0.49,0.74}
\usepackage[pagebackref,breaklinks,colorlinks,allcolors=cvprblue]{hyperref}

\def\paperID{1770} 
\def\confName{CVPR}
\def\confYear{2025}

\title{Omnidirectional Multi-Object Tracking}

\author{
Kai Luo$^{1,}$\thanks{Equal contribution} \quad Hao Shi$^{2,*}$ \quad Sheng Wu$^{1}$ \quad Fei Teng$^{1}$ \quad Mengfei Duan$^{1}$ \quad Chang Huang$^{1}$\\Yuhang Wang$^{1}$ \quad Kaiwei Wang$^{2}$ \quad Kailun Yang$^{1,}$\thanks{Correspondence: kailun.yang@hnu.edu.cn}\\
$^{1}$Hunan University \quad $^{2}$Zhejiang University\\
}

\let\oldtwocolumn\twocolumn
\renewcommand\twocolumn[1][]{%
    \oldtwocolumn[{#1}{
    \begin{center}
    \vskip-6ex
        \centering
        \includegraphics[width=1.0\textwidth]{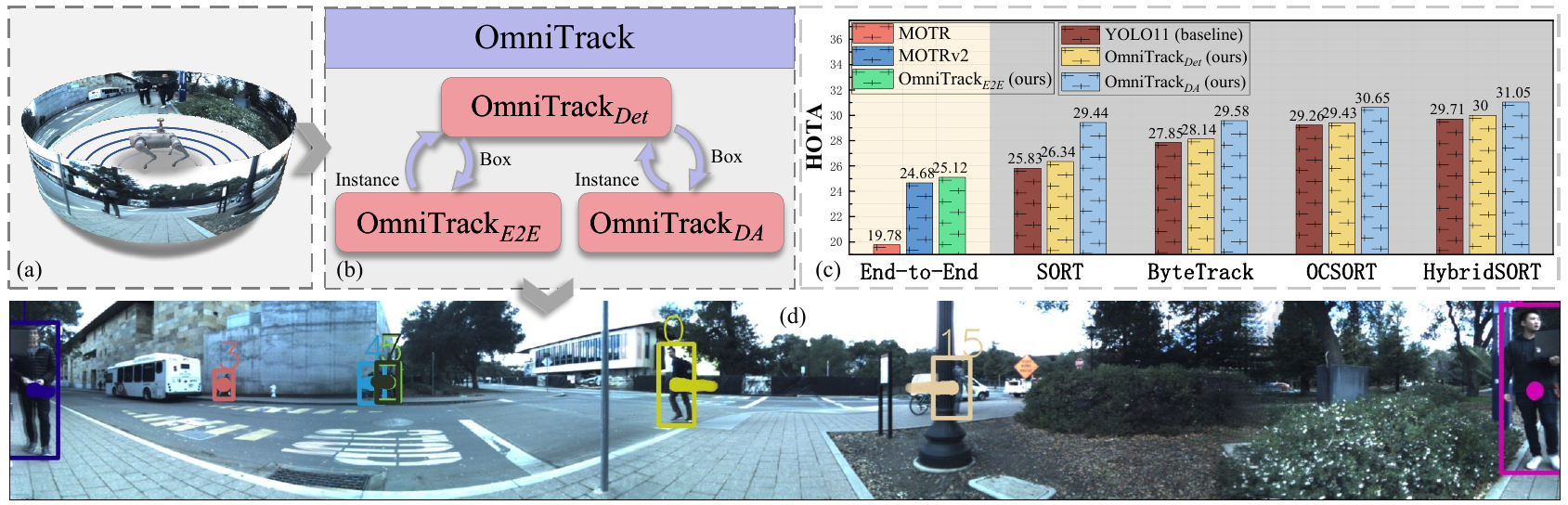}
        \vskip-2ex
        \captionof{figure} {{Comparison of OmniTrack's overall structure and performance. (a) shows the input panoramic image. (b) illustrates the proposed OmniTrack method. (c) presents a performance comparison with other multi-object tracking algorithms. (d) visualizes tracking results. 
        }}
        \label{fig:1}
    \end{center}
    }]
}

\begin{document}
\maketitle

\begin{abstract}

Panoramic imagery, with its 360{\textdegree} field of view, offers comprehensive information to support Multi-Object Tracking (MOT) in capturing spatial and temporal relationships of surrounding objects. However, most MOT algorithms are tailored for pinhole images with limited views, impairing their effectiveness in panoramic settings. Additionally, panoramic image distortions, such as resolution loss, geometric deformation, and uneven lighting, hinder direct adaptation of existing MOT methods, leading to significant performance degradation. To address these challenges, we propose OmniTrack, an omnidirectional MOT framework that incorporates Tracklet Management to introduce temporal cues, FlexiTrack Instances for object localization and association, and the CircularStatE Module to alleviate image and geometric distortions. This integration enables tracking in panoramic field-of-view scenarios, even under rapid sensor motion. To mitigate the lack of panoramic MOT datasets, we introduce the QuadTrack dataset—a comprehensive panoramic dataset collected by a quadruped robot, featuring diverse challenges such as panoramic fields of view, intense motion, and complex environments. Extensive experiments on the public JRDB dataset and the newly introduced QuadTrack benchmark demonstrate the state-of-the-art performance of the proposed framework. OmniTrack achieves a HOTA score of 26.92\% on JRDB, representing an improvement of 3.43\%, and further achieves 23.45\% on QuadTrack, surpassing the baseline by 6.81\%. The established dataset and source code are available at 
\url{https://github.com/xifen523/OmniTrack}.
\end{abstract}

\section{Introduction}
\label{sec:intro}

Panoramic cameras, with a 360{\textdegree} Field of View (FoV), capture comprehensive surrounding information, making them essential for applications like autonomous driving~\cite{Wen_2024_CVPR,cao2024occlusion}, robotic navigation~\cite{van2024visual,shi2023panoflow}, and human-computer interaction~\cite{wu2024effect,han2022panoramic_activity}. 
For small-scale mobile robots, such as quadrupedal robots, panoramic cameras are especially advantageous, allowing complete environmental awareness within a single compact setup, as illustrated in Fig.~\ref{fig:1}(a).

Despite progress in Multi-Object Tracking (MOT), panoramic MOT remains underexplored. 
Existing MOT algorithms~\cite{Chen_2024_CVPR,lv2024diffmot}, developed for pinhole cameras, struggle in panoramic settings due to inherent challenges like resolution loss, geometric distortion, and uneven color and brightness distribution when unfolded (Fig.~\ref{fig:1}~(d)). 
These challenges often lead to performance degradation when applying pinhole-based algorithms to panoramic images, limiting their effectiveness for panoramic scene perception.

To address these challenges, developing an MOT algorithm capable of comprehensive perception in panoramic images with panoramic FoV is a pressing problem. 
To this end, this paper, for the first time, proposes an omnidirectional multi-object tracking framework, \textbf{OmniTrack}, specifically designed for such tasks in 360{\textdegree} panoramic imagery. 
OmniTrack unifies two mainstream MOT paradigms—Tracking-By-Detection (TBD) and End-To-End (E2E) tracking—and introduces a feedback mechanism that effectively reduces uncertainty in panoramic FoV with rapid sensor motion, enabling fast and accurate target localization and association.

This framework consists of three core components: a \emph{CircularStatE Module}, \emph{FlexiTrack Instance}, and \emph{Tracklets Management}. 
The CircularStatE Module is designed to mitigate wide-angle distortion and enhance consistency in lighting and color.
The FlexiTrack Instance exploits the temporal continuity of objects, guiding the perception module to focus on key areas within the panoramic FoV and aiding in localization and association. 
This approach helps mitigate the difficulty of object localization in panoramic FoV. 
The Tracklets Management module collects and manages trajectory data, providing prior knowledge to the FlexiTrack Instance. 
Through these components, OmniTrack unifies the two MOT paradigms: disabling data association within Tracklets Management results in an End-to-End tracker, OmniTrack$_{E2E}$, while enabling association yields a TBD-style tracker, OmniTrack$_{DA}$. 
By employing the same data association strategy, as shown in Fig.~\ref{fig:1} (c), the framework of OmniTrack$_{DA}$ achieves significantly stronger performance. 
Disabling both the FlexiTrack Instance and Tracklets Management reduces the system to a panoramic object detector, OmniTrack$_{Det}$, as shown in Fig.~\ref{fig:1}~(b).

Moreover, to support panoramic MOT research, we developed \textbf{QuadTrack}, a dataset collected with a $360^{\circ}{\times}70^{\circ}$ panoramic camera mounted on a quadrupedal robot. This mobile platform’s biomimetic gait introduces realistic, complex motion characteristics, challenging existing MOT algorithms. Collected across five campuses in two cities, QuadTrack includes $19,200$ images, encompassing a wide variety of dynamic, real-world scenarios. In contrast to typical MOT datasets~\cite{milan2016mot16, caesar2020nuscenes,dendorfer2020mot20,semantickitti,bdd100k,cui2023sportsmot} that use static or linearly moving platforms, QuadTrack provides a new benchmark for evaluating MOT performance in panoramic-FoV scenarios with rapid and non-linear sensor motion.

At a glance, our work makes the following contributions:
\begin{itemize}
    \item To address the gap in omnidirectional multi-object tracking, we propose \textbf{OmniTrack}, a novel framework that unifies both E2E and TBD tracking paradigms. This approach reduces uncertainty and enhances perceptual and association performance in panoramic-FoV scenarios.

    \item We present \textbf{QuadTrack}, a new panoramic MOT dataset with complex motion dynamics, providing a challenging benchmark for panoramic-FoV multi-object tracking.

    \item Extensive experiments on JRDB and QuadTrack datasets show OmniTrack’s superior performance, achieving a $26.92\%$ HOTA on JRDB and $23.45\%$ on QuadTrack test splits, advancing the state-of-the-art in panoramic MOT.
\end{itemize}

\section{Related Work}
\label{sec:related_works}

\noindent\textbf{Panoramic scene understanding.}
Panoramic perception enables a holistic understanding of a 360{\textdegree} scene in a single shot~\cite{gao2022review,chen2024360+,dong2024panocontext,ehsanpour2022jrdb_act,jiang2024minimalist,jiang2022annular,ai2022deep}. 
Main areas include 
panoramic scene segmentation~\cite{teng2024360bev,zheng2024360sfuda++,cao2024occlusion,zheng2024semantics,yan2023panovos,jaus2021panoramic_panoptic,jaus2023panoramic_insights}, 
panoramic estimation~\cite{bai2024glpanodepth,ai2024elite360d,wang2022bifuse++,shen2022panoformer,chang2023depth_neural}, 
panoramic layout estimation~\cite{yu2023panelnet,shen2023disentangling,ling2023panoswin}, 
panoramic generation~\cite{zhou2025dreamscene360,wang2024360dvd,li2023panogen}, 
and panoramic flow estimation~\cite{shi2023panoflow,li2022deep}, \etc~\cite{park2024fully,kim2024fully,fan2024learned,han2022panoramic_activity}.
Researchers typically unfold panoramas into equirectangular projections or polyhedral projections to adapt algorithms designed for limited-FoV data~\cite{jiang2021unifuse,wang2022bifuse++,li2022deep}. 
They also apply techniques such as deformable convolutions to handle severe distortions in high-latitude regions~\cite{shi2023panoflow,zhang2024behind}.

Recently, researchers have recognized the advantages of omnidirectional images for tracking, particularly their ability to maintain continuous observation of targets without the out-of-view issues present in limited field-of-view setups.
Jiang~\etal~\cite{jiang2021500} propose a $500$FPS omnidirectional tracking system using a three-axis active vision mechanism to capture fast-moving objects in complex environments.
The 360VOT benchmark~\cite{huang2023360vot} is introduced for omnidirectional object tracking, focusing on spherical distortions and object localization challenges.
Huang~\etal~\cite{huang2024360loc} present 360Loc for omnidirectional localization that tackles cross-device challenges by generating lower-FoV query frames from 360{\textdegree} data. 
Another work by Xu~\etal~\cite{xu2024360vots} introduces an extended bounding FoV (eBFoV) representation to alleviate spherical distortions in panoramic videos.
Unlike previous methods, this work first explores extremely challenging panoramic-FoV and intense-motion panoramic tracking for mobile robots, \eg, aiming to enhance the robot’s spatiotemporal understanding of objects in its surroundings.

\noindent\textbf{Multi-object tracking.}
Object tracking primarily follows two paradigms: Tracking-By-Detection (TBD)~\cite{Chen_2024_CVPR,Du_2024_CVPR,qin2024towards,nettrack2024cvpr,huang2024deconfusetrack,lv2024diffmot,li2023ovtrack,qin2023motiontrack} and End-To-End (E2E)~\cite{ding2024adatrack,li2023end,MeMOTR,zeng2022motr}. 
Among these, TBD is currently one of the most prevalent, with frameworks following the design principles of SORT~\cite{wojke2017simple}. 
First, the detection network~\cite{yolox2021,carion2020end} is used to locate bounding boxes for objects, then the target's current position is predicted based on its historical trajectory, and the predicted results are associated with detection results~\cite{kuhn1955hungarian}. 
Many subsequent works have refined this approach: DeepSORT~\cite{li2022deep} introduced a ReID model to incorporate appearance information for association, and ByteTrack~\cite{zhang2022bytetrack} designed a confidence-based, stage-wise association strategy. 
Other methods~\cite{aharon2022bot,yi2024ucmc,du2023strongsort} introduced motion compensation modules to mitigate camera motion, and OC-SORT~\cite{cao2023observation} optimized the motion estimation module. Additionally, E2E methods have continued to evolve. 
TrackFormer~\cite{meinhardt2021trackformer} and MOTR~\cite{zeng2022motr} proposed transformer-based, End-to-End tracking approaches. 
Recent improvements~\cite{zhang2023motrv2, lv2024diffmot} have enhanced detector performance and improved data association accuracy in occlusion scenarios. 
Unlike existing methods that focus on narrow-FoV pinhole camera data with linear sensor motion, we address the challenges of MOT in panoramic-FoV scenarios, tackling issues such as geometric distortion and complex motion.
\begin{figure}[!t]
  \centering
  \includegraphics[width=0.48\textwidth]{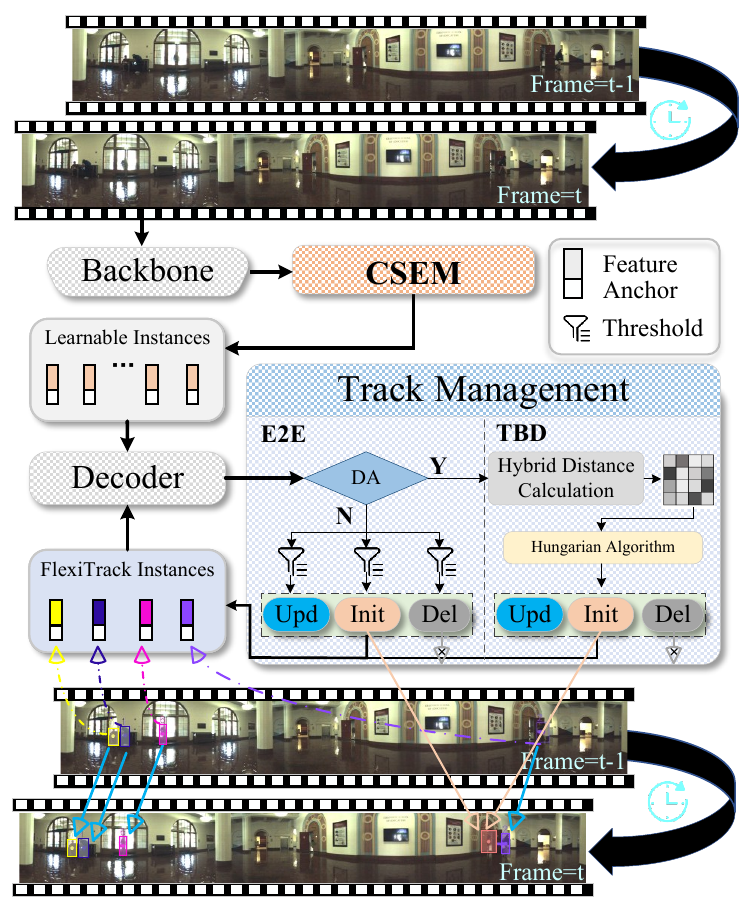}
  \vskip -1ex
  \caption{The proposed OmniTrack pipeline. \textbf{CSEM} refers to the CircularStatE Module \ref{subsec:CSEM}
, \textbf{DA} stands for data association, \textbf{E2E} denotes the End-to-End tracking paradigm, \textbf{TBD} refers to the Track-By-Detection tracking paradigm, \textbf{Upd} refers to updating tracks, \textbf{Init} to initializing tracks, and \textbf{Del} to deleting tracks.}

  \label{fig:pipeline}
  \vskip -2ex
\end{figure}

\begin{algorithm}
\caption{OmniTrack Inference Process}
\algorithmfootnote{In \textcolor{codegreen}{green} is the key of our method. }

\label{alg:Omnitrack}
\footnotesize
\KwIn{
A Panoramic video/image sequence $\texttt{V}$
}

\KwOut{
Tracks $\mathcal{T}$ of the video/image sequence
}

Initialization: $\mathcal{T} \leftarrow \emptyset$\;

Define the Initialize threshold $\mathcal{\tau}_\mathcal{I}$ \;
Define the Update threshold  $\mathcal{\tau}_\mathcal{U}$ \;

\For{frame $f_k$ in $\texttt{V}$}{
\tcc{As shown in Fig. \ref{fig:pipeline}}

$\{\mathcal{S}_3, \mathcal{S}_4, \mathcal{S}_5\} \leftarrow \texttt{Backbone}(f_k)$ \;

\textcolor{codegreen}{$ \mathcal{I}_L \leftarrow  \texttt{CSEM}(\{\mathcal{S}_3, \mathcal{S}_4, \mathcal{S}_5\})$} \;

\textcolor{codegreen}{$ \mathcal{I}_F \leftarrow  \mathcal{T}_{f_{k-1}}$} \;

$ \textcolor{codegreen}{\mathcal{D}_k^F}, \mathcal{D}_k^L \leftarrow  \texttt{Decoder} ( \textcolor{codegreen}{\mathcal{I}_F}, \mathcal{I}_L) $ \;

\BlankLine
\BlankLine
    \If{$\texttt{DA}$}{
        \tcc{Data Association}
	
        $\mathcal{C} \leftarrow  \texttt{Distance Calculation}(\textcolor{codegreen}{{\mathcal{D}_k^F}}+ \mathcal{D}_k^L, \mathcal{T}_{f_{k-1}})$ \;

        $\{\texttt{Update, Initialize, Delate} \} \leftarrow \texttt{Hungarian Algorithm}(\mathcal{C})$ \;
    
        $\mathcal{T}_{f_{k}} \leftarrow \{\texttt{Update, Initialize, Delete} \} $
    }
    
    \Else{
        \tcc{End-to-End}
        
            \textcolor{codegreen}{\For{$d$ in $ \{\mathcal{D}_k^F \cup \mathcal{D}_k^L$\} }{
	            \If{$ d \in \mathcal{D}_k^F \And d.score > \mathcal{\tau}_\mathcal{U} $}{
	               $\texttt{Update} \leftarrow  d$ \;
                }
                    \If{$ d \in \mathcal{D}_k^L \And d.score > \mathcal{\tau}_\mathcal{I}  $}{
	               $\texttt{Initialize } \leftarrow  d$ \;
                }
                \Else{
                    $\texttt{Delete} \leftarrow  d$ \;
                }
            }
        }
        $\mathcal{T}_{f_{k}} \leftarrow \{\texttt{Update, Initialize, Delete} \} $
    }
}

\textbf{Return}: $\mathcal{T}$

\end{algorithm}

\section{OmniTrack: Proposed Framework}

In this section, we introduce OmniTrack, a panoramic multi-object tracking framework that addresses the unique challenges in panoramic-FoV images, including extensive search spaces, geometric distortion, resolution loss, and lighting inconsistencies. 
OmniTrack is designed with a feedback mechanism to iteratively refine object detection, integrating trajectory information back into the detector to enhance tracking stability across panoramic-FoV scenes (Sec.~\ref{subsec:Framework}).
Specifically, we propose the OmniTrack framework, which consists of three key components:

\begin{itemize}
    \item \textbf{Tracklets Management} (Sec.~\ref{subsec:Tracklets Management}): Manages object trajectory lifecycles and provides temporal priors to the perception module.
    \item \textbf{FlexiTrack Instance} (Sec.~\ref{subsec:Temporal Query}): Rapidly locates and associates objects across the panoramic view by leveraging temporal context.
    \item \textbf{CircularStatE Module} (Sec.~\ref{subsec:CSEM}): Mitigates geometric distortion and improves consistency across the panoramic FoV, enhancing feature reliability.
\end{itemize}

\subsection{Feedback Mechanism}
\label{subsec:Framework}

The OmniTrack framework, illustrated in Fig.~\ref{fig:pipeline}, incorporates a feedback mechanism that iteratively refines detections by integrating trajectory information back into the detector. This mechanism operates on the principle of reducing information entropy, thereby enhancing stability in Panoramic-FoV and improving MOT performance.

In traditional MOT~\cite{zhang2022bytetrack,cao2023observation,du2023strongsort,aharon2022bot}, detection and association are decoupled, leading to higher entropy as each frame’s detection \( H(x_t) \) is calculated independently:
\begin{align}
H(x_t) = -\sum_{i=1}^{n} P(x_t^i) \log P(x_t^i),
\end{align}
where \( x_t^i \) denotes the position of the \( i \)-th target in frame \( t \), with probability distribution \( P(x_t^i) \). The global association entropy~\(H(\{y_t\}) \) depends on the joint probability distribution of target positions across all frames:
\begin{align}
H(\{y_t\}) = -\sum_{i=1}^{n} &P(\{x_1^i, x_2^i, \dots, x_T^i\}) \notag  \\ \times  & \log P(\{x_1^i, x_2^i, \dots, x_T^i\}). 
\label{eq:2}
\end{align} 
The cumulative entropy across all frames, accounting for independent matching, is formulated as:
\begin{align}
H_{\text{independent}} = \sum_{t=1}^{T} H(x_t) + H(\{y_t\}).
\end{align}
In contrast, OmniTrack’s feedback mechanism allows detections from frame \( t{-}1 \) to inform those in frame \( t \), reducing per-frame uncertainty. Specifically, the conditional entropy of frame \( t \), given prior feedback \( y_{t-1} \), is:
\begin{align}
H(x_t | y_{t-1}) = -\sum_{i=1}^{n} P(x_t^i | y_{t-1}^i) \log P(x_t^i | y_{t-1}^i).
\end{align}
The total entropy with feedback becomes:
\begin{align}
H_{\text{feedback}} = \sum_{t=1}^{T} H(x_t | y_{t-1}),
\end{align}
where \( H_{\text{feedback}} {<} H_{\text{independent}} \), indicating a reduction in uncertainty over time. This feedback-driven approach thus enhances tracking stability in panoramic-FoV scenarios.

\label{subsec:Mamba encoder}
\begin{figure}[!t]
  \centering
  \includegraphics[width=0.48\textwidth]{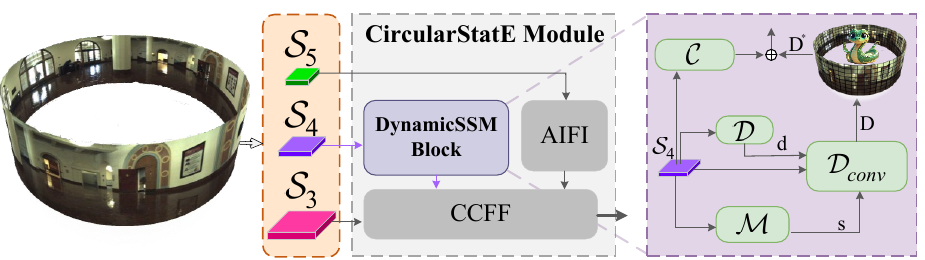}
  \caption{The proposed \textbf{CircularStatE Module} fuses multi-scale features to generate learnable instances. The \textbf{DynamicSSM Block} mitigates distortions in panoramic-FoV images, enhancing feature stability across uneven lighting and color distributions.}
  \label{fig: CircularStatE Module}
  \vskip -2ex
\end{figure} 

\subsection{Tracklets Management}
\label{subsec:Tracklets Management}

To reduce uncertainty in target localization and association while incorporating temporal information, OmniTrack incorporates a Tracklets Management module. 
During training, this module caches temporal data for instances with confidence scores exceeding a threshold \(\tau\), providing historical context to improve detection consistency in subsequent frames.
During inference, Tracklets Management oversees trajectory lifecycle management by updating, deleting, or initializing instances based on their confidence scores. In scenarios without data association, trajectories are managed directly, forming OmniTrack\(_{E2E}\) (Alg.~\ref{alg:Omnitrack}, Lines 14-21). When data association is enabled, Tracklets Management utilizes TBD-based methods~\cite{cao2023observation,yang2024hybrid} to enhance tracking, referred to as OmniTrack\(_{DA}\) (Alg.~\ref{alg:Omnitrack}, Lines 10-12)

\subsection{FlexiTrack Instance}
\label{subsec:Temporal Query}
As described in Eq.~(\ref{eq:2}), the global association entropy is significantly high under panoramic-FoV conditions, making the association task challenging. 
Benefiting from the Feedback Mechanism~(Sec.~\ref{subsec:Framework}), which integrates trajectory information into the detector to reduce information entropy.
This approach eliminates the need for global search across the entire field of view, making it especially effective for panoramic-scale perception tasks.
Based on this insight, we introduce \emph{FlexiTrack Instance}.

Each FlexiTrack Instance (see Fig.~\ref{fig:pipeline}) shares the Decoder network structure with Learnable Instances, consisting of a feature vector \( \mathcal{X} {\in} \mathbb{R}^{128} \) and an anchor \( \mathcal{Y} {\in} \mathbb{R}^{128} \), as shown in Fig.~\ref{fig:pipeline}. 
By sharing the decoder, FlexiTrack Instances can seamlessly adapt to various MOT paradigms, enhancing flexibility and allowing integration across different approaches without additional modifications.
To enhance robustness, noise is added to both feature vectors and anchors during training, minimizing dependency on historical data and improving generalization:
\begin{align}
\mathcal{X}' = \mathcal{X} + \mathcal{N}_X, \quad \mathcal{Y}' = \mathcal{Y} + \mathcal{N}_Y, 
\end{align}
where \( \mathcal{N}_X \) and \( \mathcal{N}_Y \) represent the noise components added to the feature vector and anchor, respectively.
To initialize all FlexiTrack Instances, let \( \mathcal{I_F} \) denote the set of initial \emph{instances}, and \( N \) the total number of trajectories. 
Each instance \( \mathcal{I_F}^i \) is composed of a feature vector \( \mathcal{X}_i \) and an anchor \( \mathcal{Y}_i \), as:
\begin{align}
\mathcal{I_F} = \left\{  \mathcal{I_F} ^i \mid \mathcal{I_F}^i = (\mathcal{X}'_i, \mathcal{Y}'_i), i \in \{1, 2, \dots, N\} \right\}.
\end{align}
\( \mathcal{X}'_i {\in} \mathbb{R}^{d_{\mathcal{X}}} \) and \( \mathcal{Y}'_i {\in} \mathbb{R}^{d_{\mathcal{Y}}} \) are the feature vector and anchor of the \( i \)-th trajectory, 
with \( d_{\mathcal{X}} {=} d_{\mathcal{Y}} {=} 128 \) representing their respective dimensions.
This enables $\mathcal{I_F}$ to inherit trajectory information, guiding the perception module to quickly locate the object and establish temporal associations.

\begin{table*}
	\centering
	\small
	\resizebox{0.98\textwidth}{!}{
		\setlength\tabcolsep{8pt}
		\renewcommand\arraystretch{1.0}
    \begin{tabular}{l|cc|cc|cccc}
        \topline
        \rowcolor{mygray}
          & \multicolumn{2}{c|}{\textbf{Data}} & \multicolumn{2}{c|}{\textbf{Domain}}   & &&&\\
        
        \rowcolor{mygray}
          \multirow{-2}{*}{ \textbf{Datasets}}   &  \textbf{Cov.}  & \textbf{Pano.}  & \textbf{Platform}  & \textbf{Movement}  &  \multirow{-2}{*}{\textbf{Trk Len}} & \multirow{-2}{*}
        {\textbf{No. Seq}} &\multirow{-2}{*}{\textbf{No. Smp}} &\multirow{-2}{*}{\textbf{No. T}} \\

        \hline\hline
        
 \textbf{KITTI MOT} \cite{geiger2013vision}                  &   \textit{n.a.}   & \crossmark &  \car & \wheels &  \textit{n.a.}  &  21    &   8k       &  749       \\
  \textbf{Waymo}  \cite{waymo}                   &  220$^{\circ}$ & \crossmark &  \car &  \wheels  & 20s & 103k     &      20m       &     \textit{n.a.}    \\
  \textbf{nuScenes}  \cite{caesar2020nuscenes}       &             360$^{\circ}$    & \crossmark & \car  & \wheels &  20s &   1000   &     40k  & \textit{n.a.}    \\
   \textbf{BDD100K MOT}  \cite{bdd100k}               &    \textit{n.a.}  & \crossmark &  \car  &  \wheels & 40s  &   2000   &        398k     &    \textit{n.a.}    \\
   
     \textbf{SportsMOT}   \cite{cui2023sportsmot}               &    \textit{n.a.}  & \crossmark &  \webm & \stationary    & \textit{n.a.} &   240   &      150k       &   3401      \\
     \textbf{DanceTrack}   \cite{peize2021dance}               &  \textit{n.a.}    & \crossmark &  \webm &  \stationary   & \textit{n.a.}  &  100    &      105k       & 990       \\

     \textbf{JRDB} \cite{martin2021jrdb}                  &  360$^{\circ}$    & \crossmark &  \robot &  \wheels  &  $\leq$117s &   54   &      20k       &     \textit{n.a.}    \\
    \textbf{MOT17}   \cite{milan2016mot16}              &   \textit{n.a.}   & \crossmark &  \webm & \gait \stationary   & $\leq$85s  &   14   &       11k      &  1331       \\
    \textbf{MOT20}  \cite{dendorfer2020mot20}              &  \textit{n.a.}    & \crossmark &  \webm & \stationary    & $\leq$133s &   8   &    13k     & 3833   \\
           \hline
    \rowcolor{tabgray} \textbf{QuadTrack (ours)}                &   360°    &  \mycheckmark  & \robotdog  &  \gait   &   60s &     32 &       19k      &    332     \\    
        \bottomline
        
    \end{tabular}
	}
 \vspace{-3mm}
	\captionsetup{font=small}
	\caption{Typical datasets for 2D tracking. Abbreviations:  \car~(Autonomous Car), \robot~(Mobile Robot), \robotdog~(Quadruped Robot), \mywebm~(Internet images/videos), \wheels~(Wheels), \gait~(Gait), \stationary~(Stationary), Cov. (Coverage), Pano. (Panoramic camera), %
 Trk Len (Track Length), No. Seq (The number of sequences), No. Smp (The number of samples), and No. T (the number of tracks).}
        \vspace{-3mm}
	\label{tab:comparison dataset}
	\vspace{-8pt}
\end{table*}

\subsection{CircularStatE Module} 
\label{subsec:CSEM}

The panoramic image provides an exceptionally panoramic FoV, capable of capturing 360{\textdegree} scenes. However, this inevitably introduces issues such as geometric distortions and inconsistencies in color and brightness in real-world high-dynamic-range scenes. To address these challenges, this paper proposes the \emph{CircularStatE Module}, which alleviates distortions and improves the consistency of image features, thereby enhancing the performance of perception models.

The \emph{DynamicSSM Block}, which is central to the \emph{CircularStatE Module}, is responsible for mitigating distortions and refining the feature map. The operation is broken down into the following steps: 
\label{mod:DynamicSSM}

\noindent \textbf{Distortion and Scale Calculation.} The first step is to compute both the {distortion} and {scale} information from the input feature map \( S_4 \):
\begin{align}
\mathbf{d}, \mathbf{s} = \mathcal{D}(S_4), \, \sigma(\mathcal{M}(S_4)),
\end{align}
where, \( \mathbf{d} \) and \( \mathbf{s} \) represent the distortion and scale, respectively, both of which have dimensions \( \mathbb{R}^{B \times C \times W \times H} \).

\noindent \textbf{Mitigate Distortion.} To correct distortions, we apply a dynamic convolution \( \mathcal{D}_{conv} \) to refine the feature map. The operation can be expressed as:
\begin{align}
\mathbf{D} = \mathcal{D}_{conv}(\mathbf{d} \odot \mathbf{s}, S_4),
\end{align}
where the symbol $\odot$ represents the Hadamard product, ensuring effective integration of scale adjustments.

\noindent \textbf{Improve Consistency.}
Following distortion correction, a State Space Model (SSM) \cite{mamba2} is applied to enhance light and color consistency in the panoramic image. The input to this step is the output from the previous stage, denoted as \(\mathbf{D}{\in}\mathbb{R}^{B\times C\times W\times H}\), and can be represented as follows:
\begin{align}
{\mathbf{D^\ast}}[b,c,x,y]=\frac{1}{N}\sum_{d\in\{scan\}}F_{S6}(S_d(\mathbf{D}[b,c,x,y])),
\end{align}
where \(N\) represents the number of scans, \(S_d\) represents the scanning function, and \(F_{S6}\) is the transformation function for the S6 block \cite{mamba2}.

\noindent \textbf{Feature Fusion.}
Finally, the outputs from the dynamic convolution branch and the residual branch are fused. The fusion module \( \mathcal{F} \) combines the refined feature map \( {\mathbf{D^\ast}}\) with a processed version of \( S_4 \) (obtained via a CNN operation \( \mathcal{C}(S_4) \)) to yield the final output feature map \( \mathbf{F} \):
\begin{align}
\mathbf{F} = \mathcal{F}(\mathcal{C}(S_4) \oplus {\mathbf{D^\ast}}).
\end{align}
\( \oplus \) denotes the feature fusion operation, combining details from both branches for optimal feature representation.

\section{QuadTrack: a Dynamic 360{\textdegree} MOT Dataset}
\label{sec:QuadTrack}

Most existing MOT datasets~\cite{milan2016mot16,dendorfer2020mot20,peize2021dance} are captured using pinhole cameras, which are characterized by a narrow-FoV and linear sensor motion.
However, when panoramic-FoV capture devices experience even slight movements, the entire scene can change drastically, posing significant challenges for object tracking.
QuadTrack addresses this challenge by providing a benchmark specifically designed to test MOT algorithms under dynamic, non-linear motion conditions. 
It enables evaluating algorithm robustness in tracking objects with panoramic, non-uniform motion.

\subsection{Dataset Collection and Challenges}
\begin{figure}[!t]
  \centering
  \includegraphics[width=0.48\textwidth]{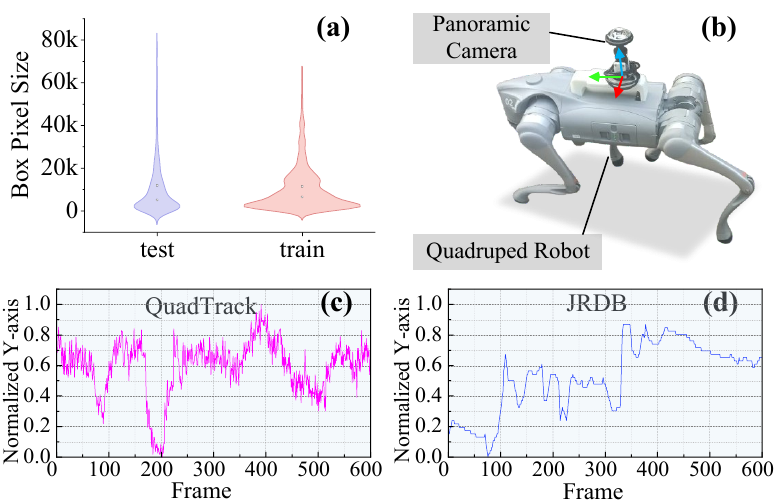}
  \vskip -2ex
  \caption{(a) shows the bounding box (bbox) size distribution for the training and validation sets, whereas (b) depicts the data collection platform and panoramic camera setup. (c) and (d) compare the normalized Y-axis pixel positions of trajectories between the QuadTrack~(\robotdog) and JRDB~\cite{martin2021jrdb}~(\robot) datasets, illustrating the significant vertical motion of the sensor in QuadTrack.}
  \label{fig:robot_platform}
  \vskip -2ex
\end{figure}

To acquire a dataset with a panoramic FoV and complex motion dynamics, we utilized a quadruped robot dog as the data collection platform. This platform was selected for its biomimetic gait, which emulates the natural locomotion patterns of quadrupedal animals, introducing additional challenges for motion tracking due to its inherent complexity and variability. 
The robot measures $70cm{\times}31cm{\times}40cm$, with a maximum payload capacity of $7kg$. It can navigate vertical obstacles up to $15cm$ and inclines up to $30^{\circ}$, making it highly maneuverable in everyday environments. 
With $12$ joint motors, the robot replicates realistic walking motions at speeds up to $2.5m/s$.
For sensing, we used a Panoramic Annular Lens (PAL) camera to capture wide-angle scenes with a FoV of $360^\circ{\times}70^{\circ}$. 
The camera has a pixel size of $3.45{\mu}m{\times}3.45{\mu}m$, a resolution of $5$ million effective pixels, and supports a maximum output of $2048{\times}2048$ pixels at $40.5$FPS. 
Mounted on the quadruped robot (see Fig.~\ref{fig:robot_platform}~(b)), the camera ensures an unobstructed, optimal view. 
Using this platform, the outdoor data collection spans morning, noon, afternoon, and evening, in diverse unconstrained environments across five campuses in two cities.

With the biomimetic gait of the quadruped robot, the collected panoramic images naturally exhibited characteristic shaking, particularly along the Y-axis (Fig.~\ref{fig:robot_platform} (c) and (d)).
Compared to the JRDB dataset~\cite{martin2021jrdb}, our QuadTrack dataset introduces more complex motion challenges. Additionally, the data faces challenges such as uneven exposure, color inconsistencies due to the panoramic FoV, and increased motion blur, as rapid relative displacement between moving objects and the background intensifies the blurring effect. 
More details can be found in the supplementary.

\subsection{Data Distribution and Comparative Analysis}
Unlike existing panoramic MOT datasets \cite{milan2016mot16,dendorfer2020mot20,geiger2013vision}, which rely on pinhole cameras, QuadTrack, as shown in Tab.~\ref{tab:comparison dataset}, is the first to be captured using a single $360^{\circ}$ panoramic camera. With a panoramic FoV ($360^{\circ}{\times}70^{\circ}$), QuadTrack significantly differs from traditional MOT datasets \cite{milan2016mot16,dendorfer2020mot20}. In contrast to autonomous driving datasets \cite{caesar2020nuscenes,bdd100k,waymo}, which often feature more predictable motion, QuadTrack incorporates complex, biologically inspired gait movements. Moreover, unlike internet-sourced datasets \cite{peize2021dance,cui2023sportsmot}, QuadTrack is designed to better reflect real-world application scenarios.
While many existing datasets~\cite{vipseg,waymo,caesar2020nuscenes,bdd100k} consist of short video sequences, QuadTrack emphasizes long-term tracking, with each video lasting $60$ seconds. To further challenge data association, we downsampled the dataset to $10$FPS, resulting in $600$ frames per sequence, spread across $32$ sequences. In total, QuadTrack includes $19,200$ frames and $189,876$ bounding boxes. 

As illustrated in Fig.~\ref{fig:robot_platform} (a), the distribution of both the training and test sets is consistent, ensuring a reliable and balanced evaluation of MOT methods. This similarity in the distribution between the sets reduces the potential for bias and allows for a more accurate comparison of model performance across varying conditions. The trajectories depicted in Fig.~\ref{fig:robot_platform} (c) and (d) highlight the increased complexity of multi-object tracking under panoramic FoV conditions. Notably, the motion along the Y-axis is significantly more intense compared to JRDB~\cite{martin2021jrdb}, further increasing the difficulty of object detection and association.

\section{Experiments }
\label{sec:Experiments}

\subsection{Experiment Setup}
\begin{table}[t!]
    \centering
    \setlength{\tabcolsep}{4pt}
    \resizebox{\columnwidth}{!}{%
    \begin{tabular}{c|l|cccc}
        \topline

       \rowcolor{mygray} 
       
       &  Method   & {HOTA}$\uparrow$ & {OSPA}$\downarrow$ & {IDF1}$\uparrow$ & {MOTA} $\uparrow$  \\
        \hline

       \parbox[t]{3mm}{\multirow{3}{*}{\rotatebox[origin=c]{90}{E2E}}} & TrackFormer~\cite{meinhardt2021trackformer} &  19.16 &  0.95  & 19.66 & 17.79  \\
        & MOTRv2~\cite{zhang2023motrv2}   &  18.22 & \textbf{0.93}  & 19.30 & 12.30  \\
        & \cellcolor{tabgray}\name{}$_{E2E}$ \text{(ours)} & \cellcolor{tabgray}\textbf{21.56} & \cellcolor{tabgray}0.94 & \cellcolor{tabgray}\textbf{22.87} & \cellcolor{tabgray}\textbf{25.01} \\

        \hline
        \parbox[t]{3mm}{\multirow{8}{*}{\rotatebox[origin=c]{90}{TBD}}} 
        & SORT~\cite{bewley2016simple}    &  23.49 & 0.90  & 26.11 & 24.59  \\
        & DeepSORT~\cite{wojke2017simple}  &  22.15 &  0.95 & 23.46 & 24.88 \\
        & ByteTrack~\cite{zhang2022bytetrack}   &   25.00 &  0.86  & 27.95 & 26.59 \\
        & Bot-SORT~\cite{aharon2022bot} &    22.90  & 0.91  & 24.27 & 23.08  \\
        & OC-SORT~\cite{cao2023observation} &     25.04 & \textbf{0.84}  & 27.89 & 25.64  \\
        & HybridSORT~\cite{yang2024hybrid}  &  25.01  &  0.85 & 27.82  & 25.03  \\
        & DiffMOT~\cite{lv2024diffmot} &    19.96  & 0.95  & 20.26 &  20.05  \\
        & \cellcolor{tabgray}\name{}$_{DA}$ \text{(ours)} & \cellcolor{tabgray}\textbf{26.92} & \cellcolor{tabgray}\textbf{0.84} & \cellcolor{tabgray}\textbf{30.26} & \cellcolor{tabgray}\textbf{26.60}  \\
        \bottomline
    \end{tabular}
    }
    \vspace{-1ex}
    \caption{Comparison with state-of-the-art methods on the JRDB test set~\cite{martin2021jrdb}.}
    \vspace{-2ex}
    \label{tab:sota JRDB}
\end{table}
\vspace{ 1 ex}
\begin{table}[t!]
    \centering
    \setlength{\tabcolsep}{4pt}
    \resizebox{\columnwidth}{!}{%
    \begin{tabular}{c|l|cccc}
        \topline

        \rowcolor{mygray} &  Method  & {HOTA}$\uparrow$ & {OSPA}$\downarrow$ & {IDF1}$\uparrow$ & {MOTA} $\uparrow$  \\
        \hline

       \parbox[t]{3mm}{\multirow{3}{*}{\rotatebox[origin=c]{90}{E2E}}} & TrackFormer~\cite{meinhardt2021trackformer} &  19.62 & 0.97  & 17.75 &  \textbf{3.16}  \\
        & MOTRv2~\cite{zhang2023motrv2}    & 16.42 & \textbf{0.96}   & 17.08  & -0.06\\
        & \cellcolor{tabgray}\name{}$_{E2E}$ \text{(ours)} & \cellcolor{tabgray}\textbf{19.87} & \cellcolor{tabgray}0.98 & \cellcolor{tabgray}\textbf{19.47} & \cellcolor{tabgray}-5.89  \\


        \hline
        \parbox[t]{3mm}{\multirow{8}{*}{\rotatebox[origin=c]{90}{TBD}}} 
        & SORT~\cite{bewley2016simple}    & 14.57  & 0.98  & 15.60 & 4.81  \\
        & DeepSORT~\cite{wojke2017simple} & 21.16  & 0.96  & 22.56 & 5.12  \\
        & ByteTrack~\cite{zhang2022bytetrack}   & 20.66  & \textbf{0.94}  & 22.56 & 8.68  \\
        & Bot-SORT~\cite{aharon2022bot} & 15.77  & 0.99  & 15.65 & 5.92  \\
        & OC-SORT~\cite{cao2023observation}&  20.83 & \textbf{0.94} & 22.60 & 7.65  \\ 	 	 	
        & HybridSORT~\cite{yang2024hybrid}  &  16.64  & 0.96  & 17.38 & 6.79  \\
        & DiffMOT~\cite{lv2024diffmot} &  16.40 & 0.97  & 16.62 & 6.21 \\ 	 	 	 

        & \cellcolor{tabgray}\name{}$_{DA}$ \text{(ours)} & \cellcolor{tabgray}\textbf{23.45} & \cellcolor{tabgray}\textbf{0.94} & \cellcolor{tabgray}\textbf{26.41} & \cellcolor{tabgray}\textbf{9.68} \\

        \bottomline
    \end{tabular}
    }

    \vspace{-1ex}
    \caption{Comparison with state-of-the-art methods on the QuadTrack test set.}
    \vspace{-3mm}
    \label{tab:sota QaudTrack}
\end{table}

\begin{table*}[t!]
    \centering
    \setlength{\tabcolsep}{4pt}
    \resizebox{2\columnwidth}{!}{%
    \begin{tabular}{c|l|l|ll|lllll}
        \topline
          \rowcolor{mygray} & Association Method & Detection Method& HOTA $ \uparrow$ & IDF1$\uparrow$ & OSPA $\downarrow$  & MOTA$\uparrow$ & DetA $\uparrow$ & AssA $\uparrow$ & FPS $\uparrow$ \\
         \hline
        \parbox[t]{3mm}{\multirow{12}{*}{\rotatebox[origin=c]{90}{Track-By- Detection (TBD)}}} 
         & \multirow{3}{*}{SORT~\cite{bewley2016simple}} & YOLO11~\cite{yolo11} (baseline) & 25.83 & 29.56 & 0.915 & 31.02 & 27.62 & 24.51  & 49.18\\
         
         &  & OmniTrack$_{Det}$~\text{(ours)} & 26.34 \textcolor{codegreen}{(+0.51)} & 31.11 \textcolor{codegreen}{(+1.55)} & 0.907 \textcolor{codegreen}{(-0.008)} & 34.21 \textcolor{codegreen}{(+3.19)} & 30.52 \textcolor{codegreen}{(+2.90)} & 22.96 \textcolor{black}{(-1.55)} & 12.14 \\

        &  & OmniTrack$_{DA}$~\text{(ours)} & 29.44 \textcolor{codegreen}{(+3.10)} & 33.27 \textcolor{codegreen}{(+2.16)} & 0.927\textcolor{black}{(+0.020)} & 33.44 \textcolor{black}{(-0.77)} & 35.16 \textcolor{codegreen}{(+4.64)} & 25.06 \textcolor{codegreen}{(+2.10)} & 11.78\\
         \cline{2-10}

        & \multirow{3}{*}{ByteTrack~\cite{zhang2022bytetrack}} & YOLO11~\cite{yolo11} (baseline) & 27.85 & 32.20 & 0.896 & 34.46 & 31.49 & 25.15 & 50.36\\
         &  & OmniTrack$_{Det}$\text{(ours)} & 28.14 \textcolor{codegreen}{(+0.29)} & 32.97 \textcolor{codegreen}{(+0.77)} & 0.870 \textcolor{codegreen}{(-0.026)} & 37.36 \textcolor{codegreen}{(+2.90)} & 32.94 \textcolor{codegreen}{(+1.45)} & 24.29 \textcolor{black}{(-0.86)} & 12.24\\
         &  & OmniTrack$_{DA}$~\text{(ours)} & 29.58 \textcolor{codegreen}{(+1.44)} & 34.54 \textcolor{codegreen}{(+1.57)} & 0.859 \textcolor{codegreen}{(-0.011)} & 38.14 \textcolor{codegreen}{(+0.78)} & 34.71 \textcolor{codegreen}{(+1.77)} & 25.49 \textcolor{codegreen}{(+1.20)} & 11.83\\
         \cline{2-10}

        & \multirow{3}{*}{OC-SORT~\cite{cao2023observation}} & YOLO11~\cite{yolo11} (baseline) & 29.26 & 33.69 & 0.874 & 34.22 & 31.81 & 27.48 & 46.33\\
         &  & OmniTrack$_{Det}$~\text{(ours)} & 29.43 \textcolor{codegreen}{(+0.17)} & 34.11 \textcolor{codegreen}{(+0.42)} & 0.851 \textcolor{codegreen}{(-0.023)} & 38.72 \textcolor{codegreen}{(+4.50)} & 34.48 \textcolor{codegreen}{(+2.67)} & 25.39 \textcolor{black}{(-2.09)} & 11.59\\
         &  & OmniTrack$_{DA}$~\text{(ours)} & 30.65 \textcolor{codegreen}{(+1.22)} & 34.83 \textcolor{codegreen}{(+0.72)} & 0.838 \textcolor{codegreen}{(-0.013)} & 36.37 \textcolor{black}{(-2.35)} & 35.58 \textcolor{codegreen}{(+1.10)} & 26.76 \textcolor{codegreen}{(+1.37)} & 11.13\\
         \cline{2-10}

        & \multirow{3}{*}{HybridSORT~\cite{yang2024hybrid}} & YOLO11~\cite{yolo11} (baseline) & 29.71 & 34.16 & 0.877 & 34.71 & 31.70 & 28.39 & 44.34\\
         &  & OmniTrack$_{Det}$~\text{(ours)} & 30.00 \textcolor{codegreen}{(+0.29)} & 34.09 \textcolor{black}{(-0.07)} & 0.853 \textcolor{codegreen}{(-0.024)} & 32.32 \textcolor{black}{(-2.39)} & 35.02 \textcolor{codegreen}{(+3.32)} & 26.09 \textcolor{black}{(-2.30)} & 11.65\\
         &  & OmniTrack$_{DA}$~\text{(ours)} & 31.05 \textcolor{codegreen}{(+1.05)} & 36.06 \textcolor{codegreen}{(+1.97)} & 0.850 \textcolor{codegreen}{(-0.003)} & 38.13 \textcolor{codegreen}{(+5.81)} & 35.08 \textcolor{codegreen}{(+0.06)} & 27.78 \textcolor{codegreen}{(+1.69)} & 10.96\\

         \hline
         \hline

         \parbox[t]{3mm}{\multirow{4}{*}{\rotatebox[origin=c]{90}{ E2E}}} 
         & TrackFormer~\cite{meinhardt2021trackformer} & \multicolumn{1}{c|}{\centering\textit{n.a.}}  & 22.22 & 23.38 & 0.959 & 23.83 & 30.30 & 16.93 & 7.38\\
         & MOTR~\cite{zeng2022motr} & \multicolumn{1}{c|}{\centering\textit{n.a.}} & 19.78 & 23.25 & 0.928 & 25.44  & 25.51 & 15.61 & 12.73\\
         & MOTRv2~\cite{zhang2023motrv2} & \multicolumn{1}{c|}{\centering\textit{n.a.}} & 24.68 & 25.49 & \textbf{0.911} & 17.05 & 26.83 & 22.97 & \textbf{13.01}\\
         &   \name{}$_{E2E}$~\text{(ours)} &  \multicolumn{1}{c|}{\centering\textit{n.a.}}  & \textbf{25.12} & \textbf{27.42} & 0.925 & \textbf{34.99} & \textbf{33.35} & \textbf{19.17}  & 11.64\\
         \bottomline
    \end{tabular}
    }        
        
    \caption{Results on the JRDB validation set~\cite{martin2021jrdb}. The first four groups compare methods under the TBD paradigm, whereas the last group presents a comparison under the E2E paradigm. In the TBD paradigm, each method is evaluated under three detection methods: the baseline with YOLO11 \cite{yolo11} as the detector, the OmniTrack$_{Det}$ detector, and OmniTrack$_{DA}$. The \textcolor{codegreen}{numbers} represent the improvement relative to the previous line's method. The FPS metric is measured on a single RTX 3090 GPU with an image resolution of $4160\times480$.
    }
    \vspace{-3mm}
    \label{tab:baseline}
\end{table*}
\begin{table}[t!]
    \centering
    \setlength{\tabcolsep}{8pt}     
    \resizebox{\columnwidth}{!}{%
    \begin{tabular}{c|cc|cccc}
        \topline
        \rowcolor{mygray}  Exp. & \textit{I$_{dn}$} & \textit{I$_{ft}$} & HOTA$\uparrow$ & IDF1$\uparrow$ & OSPA$\downarrow$ & MOTA$\uparrow$ \\
        \hline
        
        \ding{192} & - & - &  0.01 & 0.00 & 1.00 & 0.00  \\ 	 	 	 

        \ding{193} & & \checkmark   & 3.80 & 1.91 & 0.99 & -0.01 \\
        \ding{194} &   \checkmark  & & 24.32 & 26.20 &0.93  & 29.25     \\ 	 	 	

        \hline
        \rowcolor{tabgray}\ding{195} & \checkmark & \checkmark  & \textbf{25.12} & \textbf{27.42} & \textbf{0.93} & \textbf{34.99}  \\
        \bottomline
    \end{tabular}
    }
    \vspace{-1mm}
    \caption{Analysis of FlexiTrack Instance: \(\textit{I}_{dn}\) represents an instance generated using Ground Truth (GT), whereas \(\textit{I}_{ft}\) refers to a FlexiTrack Instance.
    }
    \vspace{-3mm}
    \label{tab: FlexiTrack_instance.}
\end{table}

\paragraph{Datasets.}
We conduct experiments on two datasets: JRDB~\cite{martin2021jrdb} and QuadTrack. 
JRDB is a panoramic dataset designed for crowded human environments, comprising $10$ training sequences, $7$ validation sequences, and $27$ test sequences.
The panoramic images in this dataset are stitched using a wheeled mobile robot equipped with five pinhole cameras. It includes both outdoor and indoor scenes, characterized by significant occlusion and the presence of small objects. 
Additionally, some objects exhibit rapid relative motion to the robot, which presents substantial challenges for MOT algorithms. Detailed information regarding the QuadTrack dataset is elaborated in Sec.~\ref{sec:QuadTrack}.

\vspace{-3mm}
\paragraph{Metrics.}
We employ the CLEAR metrics~\cite{bernardin2008evaluating}, including MOTA, DetA, and AssA, alongside IDF1~\cite{ristani2016performance}, OPSA~\cite{martin2021jrdb}, and HOTA~\cite{luiten2021hota} for a comprehensive tracking performance evaluation. MOTA is primarily influenced by detector performance, IDF1 measures identity preservation, and HOTA integrates association and localization accuracy, making it increasingly pivotal for tracking assessment.

\vspace{-3mm}
\paragraph{Implementation details.}
To enable a fair comparison of various MOT algorithms, we retrained models on the JRDB dataset. 
For End-To-End (E2E) algorithms~\cite{zhang2023motrv2,meinhardt2021trackformer,zeng2022motr}, we trained using the default parameters from the source code on JRDB. 
For the MOT algorithms~\cite{zhang2022bytetrack,cao2023observation,bewley2016simple,yang2024hybrid} based on the TBD paradigm, we selected the advanced YOLO11-X \cite{yolo11} as the baseline detector for training on JRDB. Additionally, OmniTrack$_{Det}$ was obtained by masking the Track Management module after training the OmniTrack model and saving the detection results. The AdamW optimizer~\cite{kingma2014adam} was used, with the learning rate set to \(10^{-5}\). For additional experimental details, please refer to the supplementary. 

\subsection{Comparison with State of the Art}
\paragraph{Tracking on JRDB test set.} 
In Tab.~\ref{tab:sota JRDB}, we compare our OmniTrack with state-of-the-art methods on the JRDB test set. Firstly, our approach significantly outperforms existing algorithms across all tracking metrics, whether in comparison with End-to-End or TBD paradigms. Specifically, OmniTrack achieves an impressive HOTA of $21.56\%$ and an IDF1 of $22.87\%$ within the End-to-End framework, surpassing the current state-of-the-art method, MOTRv2~\cite{zhang2023motrv2}, by $3.34\%$ and $3.57\%$, respectively. Furthermore, in the TBD paradigm, even under the same detector conditions, OmniTrack outperforms the state-of-the-art HybridSORT~\cite{yang2024hybrid} by $1.91\%$ in HOTA and $2.44\%$ in IDF1, demonstrating its superior performance.

\vspace{-3mm}
\paragraph{Tracking on QuadTrack test set.}

In Tab.~\ref{tab:sota QaudTrack}, we present a comparison between OmniTrack and state-of-the-art methods on the QuadTrack test set. This dataset is particularly challenging, characterized by a panoramic FoV and rapid, non-linear sensor motion, which introduces significant complexities for traditional MOT algorithms. Despite these challenges, our method outperforms existing approaches, achieving the highest HOTA scores: $19.87\%$ for the E2E group and $23.45\%$ for the TBD group. 

\subsection{Paradigm Comparison}
\paragraph{Baseline.}
To further validate the advantages of OmniTrack, we conducted comparisons based on the TBD and E2E paradigms, as shown in Tab.~\ref{tab:baseline}.
In the TBD paradigm, we evaluated several baseline tracking algorithms \cite{bewley2016simple,zhang2022bytetrack,cao2023observation,yang2024hybrid}. Each tracking method was compared under three different detection setups: using YOLO11-X~\cite{yolo11} as the baseline detector, OmniTrack\(_{Det}\) as the detector (representing traditional TBD tracking where detection and tracking are independent), and OmniTrack\(_{DA}\) with a feedback mechanism for TBD tracking. In the E2E paradigm, we used MOTR~\cite{zeng2022motr} as the baseline for comparison.

\vspace{-3mm}
\paragraph{Result.}
In the TBD method, {OmniTrack\(_{Det}\)} consistently outperforms {YOLO11-X}~\cite{yolo11}, showing an average improvement of $0.2\%$ in HOTA and $0.6\%$ in IDF1. Despite {OmniTrack\(_{Det}\)} not having a speed advantage, it achieves notable improvements in accuracy. Furthermore, when comparing {OmniTrack\(_{Det}\)} to {OmniTrack\(_{DA}\)}, the latter shows an average increase of $1.7\%$ in HOTA and $1.4\%$ in IDF1, demonstrating the effectiveness of the feedback mechanism. In the E2E paradigm, {OmniTrack\(_{E2E}\)} achieved the best result HOTA of $25.12\%$ and IDF1 of $27.42\%$.

\subsection{Ablation Study}

\paragraph{Analysis of the FlexiTrack instance.}
Tab.~\ref{tab: FlexiTrack_instance.} compares experiments with and without denoise instances and FlexiTrack instances during the training phase. Experiments \ding{192} and \ding{193} demonstrate that FlexiTrack Instances are crucial for achieving the tracking objective. In Experiment \ding{194}, we observe that denoise instances, generated from Ground Truth (GT), significantly improve the HOTA score by providing stronger guidance. 
Experiments \ding{194} and \ding{195} further show that incorporating FlexiTrack instances after using denoise instances leads to a further improvement in the HOTA score.

\begin{table}[t!]
    \centering
    \setlength{\tabcolsep}{8pt}     
    \resizebox{\columnwidth}{!}{%
    \begin{tabular}{c|ccc|ccc}
        \topline
        \rowcolor{mygray}  Exp. & $\mathcal{S}_5$ & $\mathcal{S}_4$ & $\mathcal{S}_3$ & HOTA$\uparrow$ & IDF1$\uparrow$ & OSPA$\downarrow$ \\
        \hline
        \ding{192} & - &  -  &  -  & 23.296 & 25.496 &0.93415\\
        \ding{193} & MLP &  MLP  &  MLP  & 21.951   & 23.535   & 0.92151 \\
        
        \ding{194} & Conv &  Conv  &  Conv  & 23.565  & 25.814  & \textbf{0.90931}\\

        \ding{195} & \checkmark &   \checkmark &  \checkmark  & 24.724 & 26.886 & 0.91934 \\
        \ding{196} &  \checkmark  &    &    & 24.426 & 26.016 &0.92819  \\
        \ding{197} &   &   & \checkmark  & 24.539 & 26.506 & 0.92776  \\
        \hline
        \rowcolor{tabgray}\ding{198} &  & \checkmark &    & \textbf{25.120} & \textbf{27.423} & 0.92512  \\
        \bottomline
    \end{tabular}
    }
    \vspace{-1mm}
    \caption{Ablation study on the CircularStatE module. $S_3$, $S_4$, and $S_5$ represent multi-scale features extracted from the backbone~\cite{He_2016_CVPR}. \emph{MLP} refers to fully connected layers, \emph{Conv} to convolutional layers. The symbol $\checkmark$ indicates the use of \emph{DynamicSSM} \ref{mod:DynamicSSM}}

    \vspace{-3mm}
    \label{tab:CSEM}
\end{table}

\vspace{-3mm}
\paragraph{Analysis of the CircularStatE module.}
In Tab.~\ref{tab:CSEM}, we evaluate the effectiveness of \emph{DynamicSSM} in the \emph{CircularStatE}, comparing it with other common designs such as Conv and MLP. 
The results from experiments \ding{193}, \ding{194}, and \ding{195} demonstrate a clear advantage for {DynamicSSM}. Experiments~\ding{196}, \ding{197}, and \ding{198} further show that applying {DynamicSSM} to $S_4$ yields the best performance. where $S_5$, $S_4$, and $S_3$ impact MOT results. Since $S_4$ contains both high-level semantic and low-level geometric features, its effect is the most pronounced.

\vspace{-3mm}
\paragraph{Analysis of the initialization and update thresholds.}

\begin{figure}[!t]
  \centering
  \includegraphics[width=0.48\textwidth]{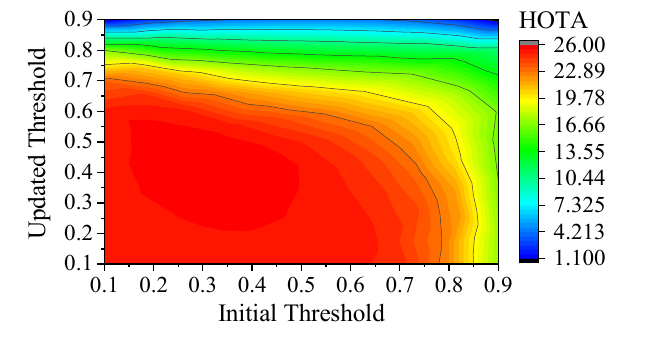}
  \vskip-3ex
  \caption{Effects of the trajectory initialization threshold and update threshold on the HOTA metric in OmniTrack$_{E2E}$.}
  \label{fig: thresholds}
  \vskip -4ex
\end{figure}

In OmniTrack\(_{E2E}\), we analyzed the impact of the \emph{initial threshold} and \emph{updated threshold} on tracking performance. As shown in Fig.~\ref{fig: thresholds}, both the \emph{initial threshold} and \emph{updated threshold} achieved HOTA scores exceeding $25\%$ within the range of $0.1$ to $0.7$. This demonstrates that OmniTrack\(_{E2E}\) is robust to threshold variations, eliminating the need for fine-tuning to achieve optimal results. 

\vspace{-3mm}
\paragraph{Comparison of end-to-end model training.}
\begin{table}[t!]
    \centering
    \setlength{\tabcolsep}{4pt}
    \resizebox{\columnwidth}{!}{
    \begin{tabular}{l|cccc}
        \topline
        \rowcolor{mygray}  Method  & {\#Params} & {FLOPs}& {MACs}  & {Training Time} $\downarrow$   \\
        \hline
        TrackFormer~\cite{meinhardt2021trackformer}   & 44.01M  & 335G   & 167G & 108 hours \\
        MOTR~\cite{zeng2022motr}  &   43.91M &  1421G & 709G &  80 hours \\
        MOTRv2~\cite{zhang2023motrv2} &  41.65M &   1395G  &  696G  &  130 hours\\
        \hline

        \rowcolor{tabgray} \name{}$_{E2E}$~\text{(ours)}    & 63.13M  &  762G   & 369G & 20 hours\\

        \bottomline
    \end{tabular}
    }
    \vspace{-1mm}
    \caption{Comparison of parameters, FLOPs, MACs, and training time for various end-to-end models on the JRDB dataset~\cite{martin2021jrdb}.}
    \vspace{-3mm}
    \label{tab:6_para}
\end{table}

In Tab.~\ref{tab:6_para}, we compare the number of parameters and training time of OmniTrack\(_{E2E}\) with existing End-to-End methods. 
Our method trains over four times faster than other End-to-End methods using default parameters on the JRDB dataset. 
This is achieved by implementing identity association through FlexiTrack Instances, which significantly simplifies the model design of the association component and alleviates the challenges associated with E2E model training.

\section{Conclusion}
\label{sec:Conclusion}
This paper presents OmniTrack, a multi-object tracking framework tailored for panoramic images, effectively addressing key challenges like geometric distortion, low resolution, and lighting inconsistencies. Central to OmniTrack is a feedback mechanism that reduces uncertainty in panoramic-FoV tracking. The framework incorporates Tracklets Management for temporal stability, FlexiTrack Instance for rapid localization and association, and the CircularStatE Module to mitigate distortion and improve visual consistency.
Additionally, we present QuadTrack, a cross-campus multi-object tracking dataset collected using a quadruped robot to support dynamic motion scenarios. 
This challenging dataset is designed to advance research in omnidirectional perception for robotics. Experiments verify that OmniTrack achieves state-of-the-art performance on public JRDB and the established QuadTrack datasets, demonstrating its effectiveness in handling panoramic tracking tasks.

\noindent\textbf{Limitations.} 
While OmniTrack demonstrates strong performance, our approach is currently limited to 2D panoramic tracking without 3D capabilities, restricting depth perception in complex scenes. Additionally, the method is centered around a mobile robotic platform. Future work could consider extending to 3D panoramic MOT or exploring human-robot collaborative perception to enhance situational awareness.

\section*{Acknowledgment}
\label{sec:Acknowledgment}
This work was supported in part by the National Natural Science Foundation of China (No.~62473139 and No.~12174341), in part by Zhejiang Provincial Natural Science Foundation of China (Grant No. LZ24F050003), and in part by Shanghai SUPREMIND Technology Co. Ltd.

{
    \small
    \bibliographystyle{ieeenat_fullname}
    \bibliography{main}
}

\clearpage

\section{Annotation of the QuadTrack Dataset}

In the annotation process of the established QuadTrack dataset, we used CVAT~\cite{cvat}, an open-source annotation tool that supports tasks such as object detection, object tracking, and instance segmentation. CVAT offers both local and online versions, providing high flexibility for users. 
Prior to annotation, we preprocessed the dataset by selecting representative scenes, including $32$ sequences (seq), with $16$ sequences allocated for training and $16$ for testing. 
Each sequence contains $600$ frames with a frame rate of approximately $10$FPS, resulting in a duration of about $60$ seconds per sequence. 
Furthermore, to assist annotators in better semantic understanding and precise labeling, we unfolded the images into a $2048{\times}480$ panoramic layout via equirectangular projection. 
For the bounding boxes at the image borders, we ensured continuous tracking, guaranteeing that the same object in the surrounding environment maintained a unique ID. The minimum bounding box area was set to $800$ pixels, and any targets smaller than this area were ignored. 
The QuadTrack dataset includes two common object classes: \emph{car} and \emph{person}.

Upon completion of the annotation process, the final annotation attributes were thoroughly reviewed and validated through a filtering and cross-validation procedure to ensure data accuracy. After ensuring the correctness of the annotations, the final annotation attributes were formatted into the MOT standard~\cite{milan2016mot16}. 
Example of ground truth:

\begin{lstlisting}[language={Python}]
# MOT format
# f_id, t_id, x, y, w, h, conf, cls, vis
# data
1,1,733.67,281.66,34.78,106.81,1,1,1.0
1,2,557.87,268.05,24.36,128.58,1,1,1.0
1,3,382.33,316.41,110.61,61.49,1,2,1.0
1,4,000.00,301.35,35.02,82.89,1,2,1.0
1,5,1917.7,278.79,20.70,97.98,1,1,1.0
...
\end{lstlisting}

\begin{figure*}[!t]
  \centering
  \includegraphics[width=0.80\textwidth]{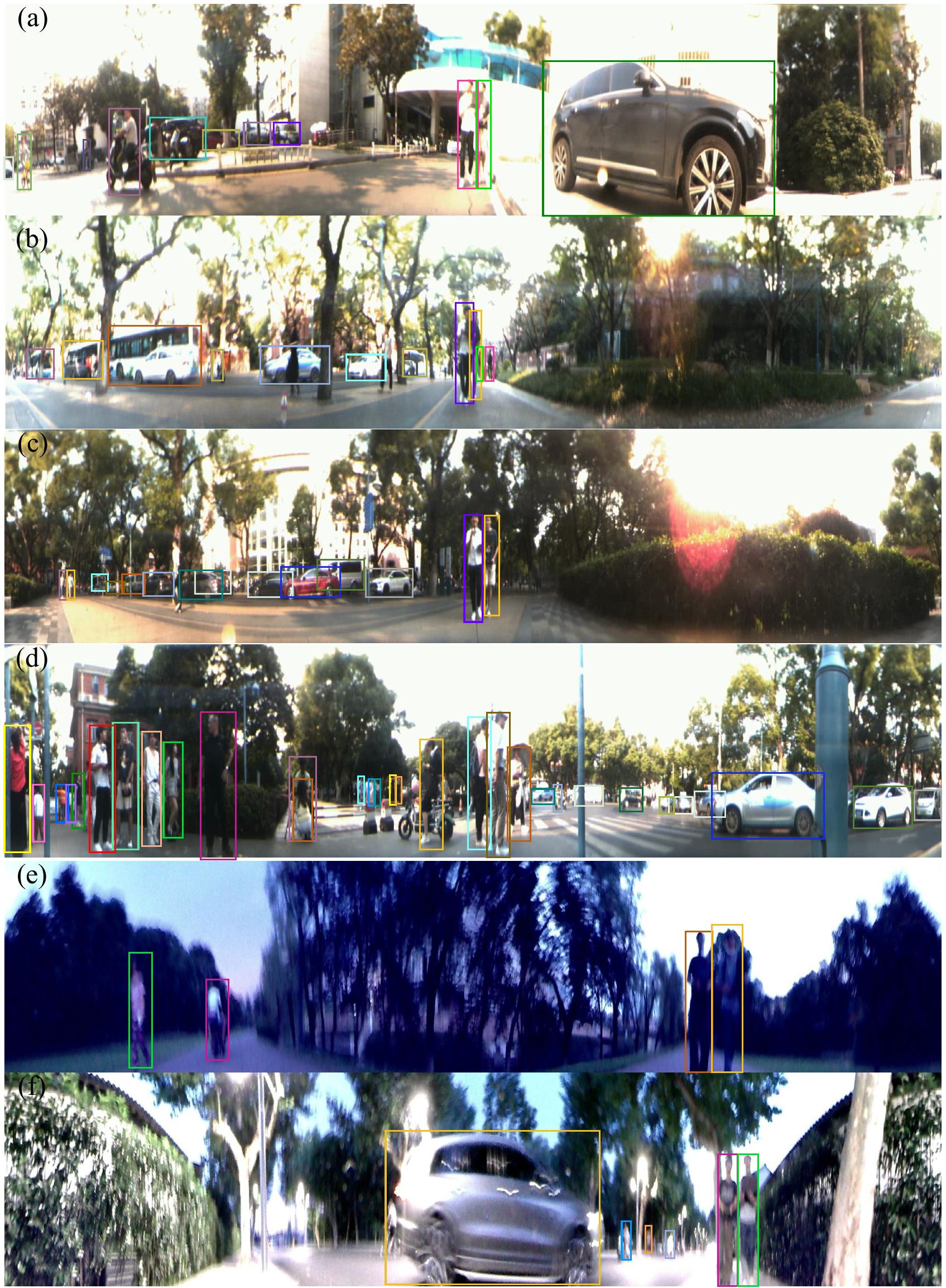}
  \caption{Examples of the established QuadTrack dataset. The QuadTrack dataset features a variety of scenes, including different campuses, streets, and low-light environments, with machine-generated labels for each scenario. These labeled scenes demonstrate the diversity and complexity of the dataset, offering insights into the challenges of multi-object tracking across different real-world contexts.}
  \label{fig:vis_anno}
\end{figure*}

For a comprehensive description of the attributes in the dataset, please refer to Tab.~\ref{tab:anno}.
This annotation format, commonly used in Multi-Object Tracking (MOT) research, provides a structured and standardized method for organizing the data. 
The inclusion of essential attributes such as object identity, bounding box coordinates, and visibility status is critical for training and assessing tracking models in dynamic, real-world environments. 
In Fig.~\ref{fig:vis_anno}, examples from the QuadTrack dataset are shown, demonstrating the diversity of scenes and the visual presentation of annotations.

\begin{figure}[!t]
  \centering
  \includegraphics[width=0.48\textwidth]{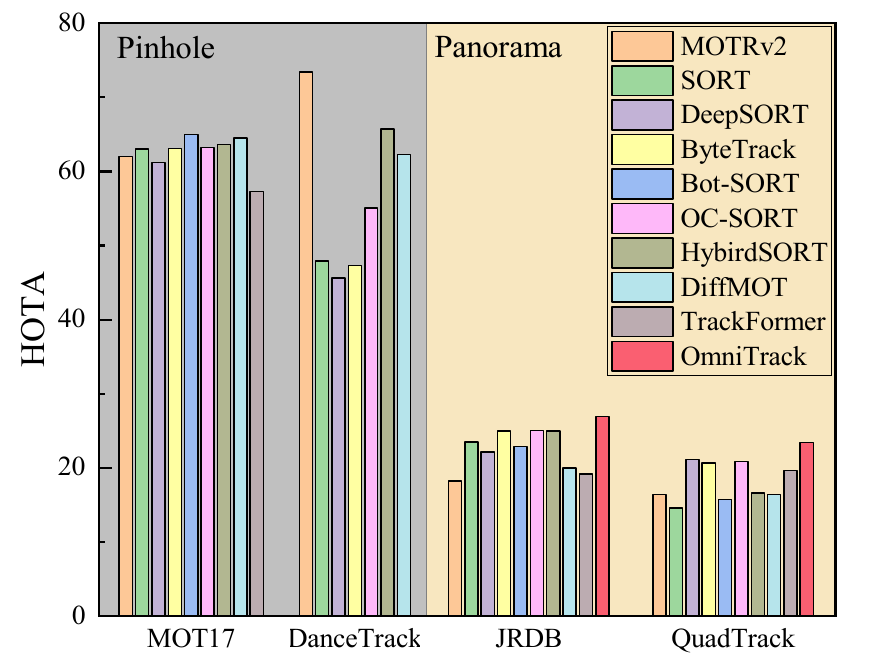}
  \caption{Comparison of state-of-the-art methods on different datasets. \textbf{Pinhole} refers to Multi-Object Tracking (MOT) datasets that utilize pinhole camera images, whereas \textbf{Panorama} refers to MOT datasets that employ panoramic images.}
  \label{fig:SOTA}
\end{figure}

\begin{table*}[t!]
    \centering
    \setlength{\tabcolsep}{16pt}     
    \small
    \begin{tabular}{lll}
        \topline
        \rowcolor{mygray}  Pos. & {Key} &  Explanation  \\
        \hline
        1 & Frame\_id & Represents the frame ID. \\
        2 & Track\_id & A unique identifier for each object. A value of -1 indicates a detection item. \\
        3 & Left & Coordinates of the top-left corner of the object bounding box. \\
        4 & Top & Coordinates of the top-left corner of the object bounding box. \\
        5 & Width & Width of the object bounding box. \\
        6 & Height & Height of the object bounding box. \\
        7 & Confidence & It acts as a flag whether the entry is to be considered (1) or ignored (0). \\
        8 & Class & Indicates the type of object annotated. \\
        9 & Visibility & Visibility ratio, a number between 0 and 1 that says how much of that object is visible.\\
        \bottomline
    \end{tabular}
    \vspace{-1mm}
    \caption{Detailed explanation of the annotation attributes for the QuadTrack dataset, including the meaning of each position.}
    \vspace{-3mm}
    \label{tab:anno}
\end{table*}

\begin{figure*}[!t]
  \centering
  \includegraphics[width=0.95\textwidth]{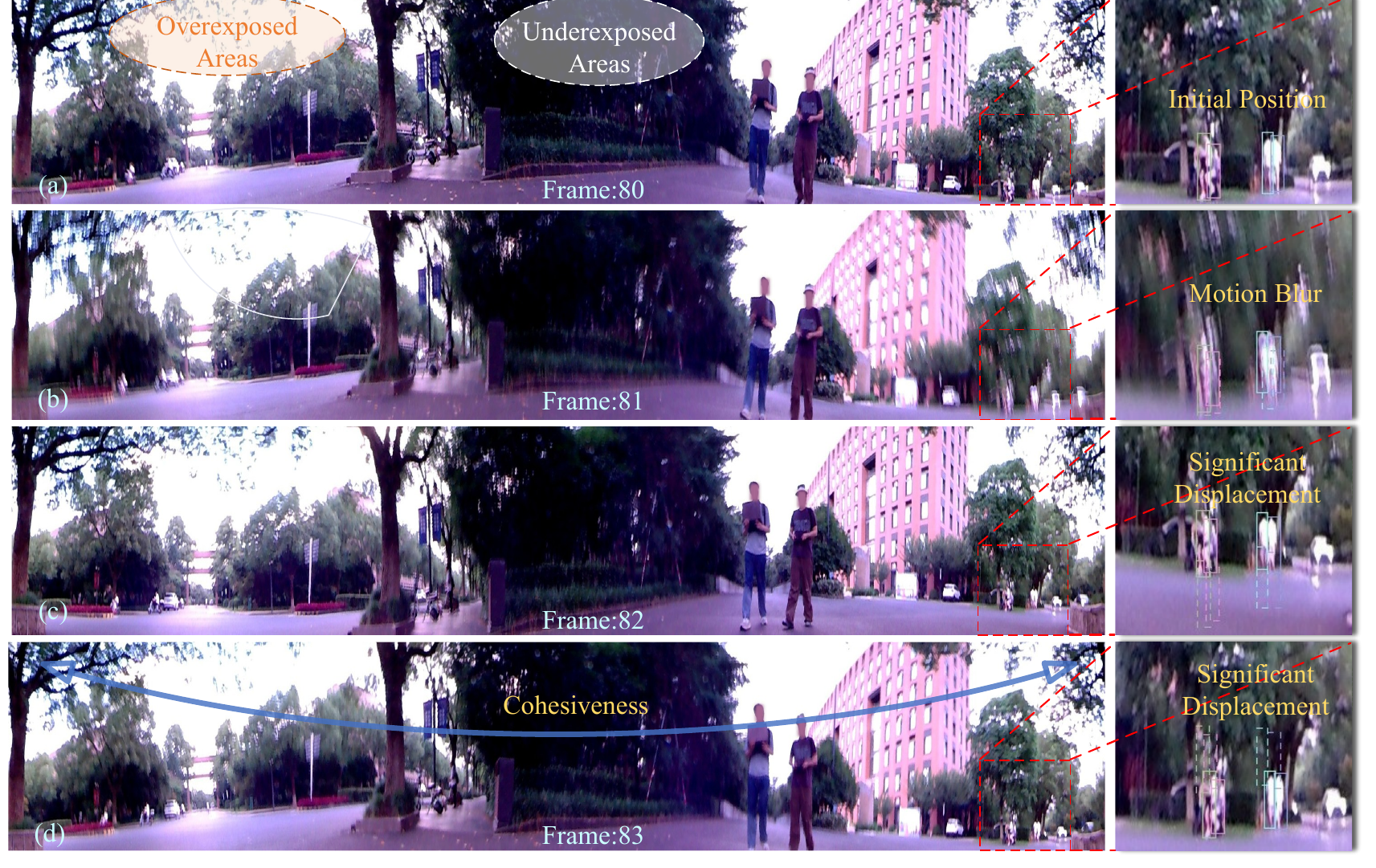}
  \caption{The QuadTrack dataset presents several significant challenges. The images labeled (a), (b), (c), and (d) illustrate continuous frames $80$ to $84$ from a sequence, with corresponding magnified views shown on the right. In these magnifications, solid rectangular boxes represent the Ground Truth (GT) for the current frame, while dashed boxes correspond to the GT from the preceding frame. One notable challenge is motion blur, particularly evident in the magnified view of frame (b), where the bionic gait introduces substantial blur to the target object. Moreover, there is considerable positional displacement between adjacent frames, as demonstrated in the magnified views of frames (c) and (d). The panoramic images also present inherent exposure issues, displaying both overexposed and underexposed regions, as seen in (a). Finally, the continuity inherent in the panoramic images presents an additional critical factor for the tracking task.}
  \label{fig:challenge}
\end{figure*}

\section{Additional Ablation Studies and Analyses}
\subsection{More Analyses of the DynamicSSM Block}
We provide a more detailed discussion on the components of the DynamicSSM Block in Tab.~\ref{tab:dssm}. 
The DynamicSSM Block is composed of three primary operations: (\romannumeral 1) distortion alleviation, as described in the main text Equation~9, (\romannumeral 2) addressing lighting and color inconsistencies, as detailed in the main text Equation~10, and (\romannumeral 3) enhancing feature representation, as formulated in the main text Equation~11. 
As shown in Tab.~\ref{tab:dssm}, all three operations individually contribute to improved performance, and their combination results in the best overall performance. 
A comparison between experiments \ding{192} and \ding{195} demonstrates that integrating all three operations in the DynamicSSM Block leads to an overall HOTA improvement of $1.82\%$.

\subsection{More Analyses of the CircularStatE Module}
In the CircularStatE Module, we designed a key component, the DynamicSSM Block, to address challenges such as distortion and lighting inconsistencies inherent in panoramic images. Compared to convolutional networks, the DynamicSSM Block offers a significant performance advantage in handling these issues. 
To further explore the impact of convolutional networks on multi-scale features, we conducted additional experiments, as summarized in Tab.~\ref{tab:conv}. 
The results show that applying a convolutional network to the S5 scale yielded the best performance for the CircularStatE Module, achieving a HOTA score of $24.107\%$.

\begin{table}[t!]
    \centering
    \setlength{\tabcolsep}{8pt}     
    \resizebox{\columnwidth}{!}{%
    \begin{tabular}{c|ccc|ccc}
        \topline
        \rowcolor{mygray}  Exp. & Dconv & SSM & Fusion & HOTA$\uparrow$ & IDF1$\uparrow$ & OSPA$\downarrow$  \\
        \hline
         	 	 
        \ding{192} & - & - & - &  23.30 & 25.50 & 0.93   \\ 	 	
        \ding{193} & - & \checkmark & \checkmark &  24.82 & 27.17 & \textbf{0.92}   \\ 	 	
        \ding{194} & \checkmark & - & \checkmark & 24.81  & 26.98 &   \textbf{0.92} \\ 
        \ding{195} & \checkmark & \checkmark & - & 24.72  & 26.66 &   \textbf{0.92} \\ 

        \hline
        \rowcolor{tabgray}\ding{195} & \checkmark & \checkmark  & \checkmark & \textbf{25.12} & \textbf{27.42} & 0.93   \\
        \bottomline
    \end{tabular}
    }
    \caption{Ablation of the DynamicSSM Block: Dconv represents deformable convolution (Equation~9 in the main text), SSM denotes the state-space model (Equation~10 in the main text), and Fusion refers to the integration of residual features (Equation~11 in the main text).
    }
    \label{tab:dssm}
\end{table}

\begin{table}[t!]
    \centering
    \setlength{\tabcolsep}{8pt}     
    \resizebox{\columnwidth}{!}{%
    \begin{tabular}{c|ccc|ccc}
        \topline
        \rowcolor{mygray}  Exp. & $\mathcal{S}_5$ & $\mathcal{S}_4$ & $\mathcal{S}_3$ & HOTA$\uparrow$ & IDF1$\uparrow$ & OSPA$\downarrow$ \\
        \hline
        \ding{192} & - &  -  &  -  & 23.296 & 25.496 &0.93415\\

        \ding{193} & Conv &  Conv  &  Conv  & 23.565  & 25.814  & \textbf{0.90931}\\

        \ding{194} &  Conv  &  -  &  -  & \textbf{24.107} & \textbf{26.374} & 0.92567 \\
        \ding{195} & -  & Conv  & - & 23.814 & 26.083  & 0.92624  \\
        \ding{196} & - & - &  Conv  & 23.721 & 25.565 & 0.91992  \\
        \bottomline
    \end{tabular}
    }
    \caption{Analysis of the impact of convolution in the CircularStatE Module. $S_3$, $S_4$, and $S_5$ represent multi-scale features extracted from the backbone~\cite{He_2016_CVPR}. \emph{Conv} represent convolution.}

    \label{tab:conv}
\end{table}

\subsection{More Analyses of Hyperparameters}

\noindent\textbf{Analysis of Impacts of Training Epochs.} We further analyzed the variations observed across different epochs by selecting the same parameters (\textit{i.e.}, track initialization threshold of $0.55$ and track update threshold of $0.45$). The experiments were conducted on the validation dataset of JRDB~\cite{martin2021jrdb}, with model weights saved every $5$ epochs, and inference was performed at the end. The results are presented in Tab.~\ref{tab:epoch}. 
As shown in the table, different epochs have a noticeable impact on the final HOTA metric. When the epoch was set to $100$, the best HOTA value of $25.12\%$ was achieved, with results from other epochs slightly lower than this value. Overall, the results demonstrate that OmniTrack exhibits strong robustness and consistent performance across different epochs.

\noindent\textbf{Analysis of FlexiTrack Instance Noise.} 
FlexiTrack Instance (Sec.~3.3 in the main text) plays a crucial role in assisting the detection module to quickly locate targets in panoramic field-of-view scenarios and establish temporal associations between them. A key aspect of its performance is the initialization phase, where the selection of motion noise can significantly influence the overall tracking results. To investigate this, we analyze the impact of different motion noise levels on FlexiTrack Instance’s performance on the validation set of JRDB~\cite{martin2021jrdb}, as presented in Tab.~\ref{tab:nosing_motion}. 
From the table, it is evident that varying motion noise levels have a notable effect on the final HOTA score. 
Specifically, a motion noise value of $0.5$ improves performance, leading to a significant boost in tracking accuracy.

\subsection{More Analyses of MOT Datasets}
To visually assess the overall performance of existing state-of-the-art methods on panoramic MOT datasets, we compare the pinhole-based MOT17~\cite{milan2016mot16} and DanceTrack~\cite{peize2021dance} datasets with the panoramic datasets JRDB~\cite{martin2021jrdb} and QuadTrack. As shown in Figure~\ref{fig:SOTA}, MOTRv2~\cite{zhang2023motrv2} achieves a HOTA of $73.4\%$ on DanceTrack~\cite{peize2021dance} but only $18.22\%$ on JRDB~\cite{martin2021jrdb}, representing a decrease of $55.18\%$. Similarly, ByteTrack~\cite{zhang2022bytetrack} achieves $63.1\%$ HOTA on MOT17~\cite{milan2016mot16} but only $20.66\%$ on QuadTrack, a drop of $42.44\%$. Overall, the HOTA on panoramic datasets is approximately $30\%$ lower than on pinhole-based datasets. More importantly, OmniTrack significantly outperforms existing SOTA methods on both panoramic datasets, marking a substantial advancement in the field of panoramic multi-object tracking.

\begin{table}[t!]
    \centering
    \setlength{\tabcolsep}{8pt}     
    \resizebox{\columnwidth}{!}{%
    \begin{tabular}{c|c|cccc}
        \topline
        \rowcolor{mygray}  Exp. &  Epoch & HOTA$\uparrow$ & IDF1$\uparrow$ & OSPA$\downarrow$ & MOTA$\uparrow$   \\
        \hline
	 
        \ding{192} & 80 &  24.16 &	25.84 &	\textbf{0.93} &	31.04           \\ 	
        
        \ding{193} & 85 &  25.05 &	27.29 &	\textbf{0.93} &	33.74           \\ 	 	 	 

        \ding{194} & 90   &       24.70 &  	26.85 	& \textbf{0.93} & 	33.09 \\ 
        \ding{195} & 95   &   24.95 &	27.31& 	\textbf{0.93} &	31.32    \\ 
        \ding{196} & 105   &24.99 &	27.25& 	\textbf{0.93} &33.05   \\ 
        \ding{197} & 110   & 25.00 &	27.20 	& \textbf{0.93} &	32.83     \\ 
        \ding{198} & 115  &  24.70& 	27.11 & 	\textbf{0.93} 	&31.75    \\

        \hline
        \rowcolor{tabgray}\ding{199} & 100  & \textbf{25.12} & \textbf{27.42} & \textbf{0.93}  & \textbf{34.99} \\
        \bottomline
    \end{tabular}
    }
    \caption{Analysis of the impact of epochs on performance. Analysis of the performance impact of the OmniTrack$_{E2E}$ method across different epochs, with other parameters held constant.
    }
    \label{tab:epoch}
\end{table}
\begin{table}[t!]
    \centering
    \setlength{\tabcolsep}{8pt}     
    \resizebox{\columnwidth}{!}{%
    \begin{tabular}{c|c|cccc}
        \topline
        \rowcolor{mygray}  Exp. &  Noise & HOTA$\uparrow$ & IDF1$\uparrow$ & OSPA$\downarrow$ & MOTA$\uparrow$   \\
        \hline
        
        \ding{192} & 0.1 & 19.72 & 20.65  & 0.95 &   28.63 \\
        \ding{194} & 0.8 &  24.32 &26.28  & \textbf{0.93}  & 34.88 \\ 
        \ding{194} & 1.0 & 23.61 & 25.84 & \textbf{0.93} &33.12 \\ 
 	 	 	 
        \hline
        \rowcolor{tabgray}\ding{195} & 0.5  & \textbf{25.12} & \textbf{27.42} & \textbf{0.93}  & \textbf{34.99} \\
        \bottomline
    \end{tabular}
    }
    \caption{Ablation of FlexiTrack Instance noise. The noise mentioned here refers to the one applied to the anchor (in Equation 6 of the main text), while the feature vector remains unchanged.
    }
    \label{tab:nosing_motion}
\end{table}
\section{Reproduction of state-of-the-art Methods.}
Due to the absence of existing performance records for SOTA methods on the JRDB and QuadTrack datasets, all comparative experiments in this paper were independently reproduced. In the reproduction process, we prioritized using the official source code, provided it was executable. The parameter selection was based on the recommendations in the original papers, aiming to achieve optimal performance on both the JRDB and QuadTrack datasets.

\subsection{Methods for the E2E Paradigm.}

\noindent\textbf{TrackFormer.}
To reproduce the TrackFormer method~\cite{meinhardt2021trackformer}, we utilized the official source code (\href{https://github.com/timmeinhardt/trackformer}{link}) and applied it to both JRDB~\cite{martin2021jrdb} and QuadTrack datasets.
Our implementation uses COCO pre-trained weights from Deformable DETR~\cite{zhu2020deformable}, incorporating iterative bounding box refinement to enhance tracking accuracy. The model is trained on a single GPU with a batch size of $2$. 
To adapt the model for JRDB~\cite{martin2021jrdb} and QuadTrack datasets, we reformat the data to align with the MOT20 format~\cite{dendorfer2020mot20}, which is a widely used format in multi-object tracking challenges. 
Training is conducted for $30$ epochs, with an initial learning rate of \( 2 {\times} 10^{-4} \). The learning rate is decayed by a factor of $10$ every $10$ epochs, as per the official guidelines.
All other parameters remain unchanged, using the default values.

\noindent\textbf{MOTR.} 
In reproducing the MOTR method~\cite{zeng2022motr}, we encountered challenges when training with the weights originally used in the TrackFormer method~\cite{meinhardt2021trackformer}. 
As a result, we opted to train the model on the JRDB dataset~\cite{martin2021jrdb} using pre-trained weights from the MOT17 dataset~\cite{milan2016mot16}, which is specifically designed for multi-object tracking tasks.
The model is fine-tuned on a single GPU with a batch size of $1$. 
To adapt the model to the JRDB dataset~\cite{martin2021jrdb}, we modified the data format to match the DanceTrack format~\cite{peize2021dance}. This format adaptation ensures compatibility with the input requirements of the MOTR framework~\cite{zeng2022motr}. The model is trained for $25$ epochs to ensure model convergence, with an initial learning rate of \( 2 {\times} 10^{-4} \). 
Based on the official source code (\href{https://github.com/megvii-research/MOTR}{link}) and our experience, the learning rate is reduced by a factor of $10$ every $5$ epochs.
All other parameters were retained at their default values, as per the official guidelines.

\noindent\textbf{MOTRv2.} 
The pre-trained weights are identical to those used in TrackFormer~\cite{meinhardt2021trackformer}. 
The model is trained on a single GPU with a batch size of $1$.
To adapt the model for JRDB~\cite{martin2021jrdb} and QuadTrack datasets, we convert the data to the DanceTrack format~\cite{peize2021dance}. 
Since MOTRv2~\cite{zhang2023motrv2} is highly dependent on detection results, we use ground truth detections for the training set to ensure optimal tracking performance. 
For the test set, to maintain fairness, we generate detection results using our own detector.
The training procedure spans $15$ epochs for JRDB~\cite{martin2021jrdb} and $25$ epochs for QuadTrack, after which the model ceases to converge. The initial learning rate is set to \( 2 {\times} 10^{-4} \) with a decay factor of $10$ every $5$ epochs, in alignment with the settings used in MOTR~\cite{zeng2022motr}.
All other parameters were retained at their default values, as specified in the official source code (\href{https://github.com/megvii-research/MOTRv2}{link}). 

\subsection{Methods for the TDB Paradigm.}

\noindent\textbf{HybridSORT.} 
In reproducing the HybridSORT method~\cite{yang2024hybrid} on both JRDB~\cite{martin2021jrdb} and QuadTrack datasets, we utilized the official source code (\href{https://github.com/ymzis69/HybridSORT}{link}). HybridSORT offers two variants: an appearance-based version and an appearance-free version. 
For all experiments presented in this paper, the appearance-free version of HybridSORT was employed. For parameter selection, consistent values were applied across both JRDB~\cite{martin2021jrdb} and QuadTrack datasets: \texttt{track\_thresh} was set to $0.6$ and \texttt{iou\_thresh} was set to $0.15$, in alignment with the settings used in the DanceTrack dataset~\cite{peize2021dance}. 
All other parameters were kept at their default values, as specified in the official implementation.

\noindent\textbf{SORT.} As a pioneering approach in the TBD paradigm, the SORT method~\cite{bewley2016simple} has multiple implementation versions. However, due to the age of the original source code, it has been deprecated. In this paper, we chose to reproduce the SORT method based on the HybridSORT~\cite{yang2024hybrid} source code \href{https://github.com/ymzis69/HybridSORT}{(link)}. 
For both JRDB~\cite{martin2021jrdb} and QuadTrack datasets, we set \texttt{track\_thresh} to $0.6$ and \texttt{iou\_thresh} to $0.3$, in alignment with the settings used for the SORT method on the DanceTrack dataset~\cite{peize2021dance}. 
All other parameters were retained at their default values, as per the official guidelines.

\noindent\textbf{DeepSORT.} In the comparative experiments of this paper, we encountered compatibility issues with the DeepSORT~\cite{wojke2017simple} source code repository, which was not compatible with Torch models, complicating the reproduction process. 
As a result, we chose to reproduce the DeepSORT algorithm using the code from HybridSORT~\cite{yang2024hybrid}. It is important to note that DeepSORT is an appearance-based tracking method, which, in theory, requires the separate training of the appearance module for both JRDB~\cite{martin2021jrdb} and QuadTrack datasets. 
However, due to the lack of explicit guidance on training the appearance weights, we used the pre-trained appearance weights provided in the source code, specifically the \texttt{googlenet\_part8\_all\_xavier\_ckpt\_56.h5} checkpoint. 
All other parameters were retained at their default values and were not modified.

\noindent\textbf{ByteTrack \& OC-SORT.} 
In reproducing ByteTrack~\cite{zhang2022bytetrack} and OC-SORT~\cite{cao2023observation}, we chose to use their official source code to ensure consistency and accuracy. All parameter settings were directly taken from the official demo configurations, which were specifically designed to optimize performance. These settings were applied uniformly across both JRDB and QuadTrack datasets to maintain a fair comparison. This approach allows for a reliable evaluation of the performance of both tracking algorithms on our datasets while adhering to the original implementation guidelines.

\noindent\textbf{BoT-SORT.} 
BoT-SORT~\cite{aharon2022bot} is a tracker in the TBD paradigm that integrates multiple techniques, including the use of appearance features. 
For both JRDB~\cite{martin2021jrdb} and QuadTrack datasets, we trained the appearance feature model using Fast-ReID~\cite{he2020fastreid}. 
All other parameters were retained as specified in the original BoT-SORT source code (\href{https://github.com/NirAharon/BoT-SORT}{link}), ensuring consistency with the default configuration. 


\subsection{YOLO11 Detection}
In the TBD paradigm of tracking, the performance heavily depends on the detector's results. 
We selected the best detector in the YOLO series~\cite{redmon2016you}, YOLO11~\cite{yolo11}, as the baseline for comparison. 
To enhance the perception capability, we selected the YOLO11 series model with the largest number of parameters, the YOLO11-X~\cite{yolo11}, for training.
The training configuration consisted of $100$ epochs, an image size of $960$, and a batch size of $8$, with all other settings maintained at their default values. Upon completion of the training, the model weights from the best-performing checkpoint, \texttt{best.pt}, were used to infer the images in the test set. Detection results with confidence scores greater than the threshold $0.1$ were retained and subsequently provided as input to the tracker in the TBD paradigm.

\begin{figure*}[!t]
  \centering
  \includegraphics[width=1\textwidth]{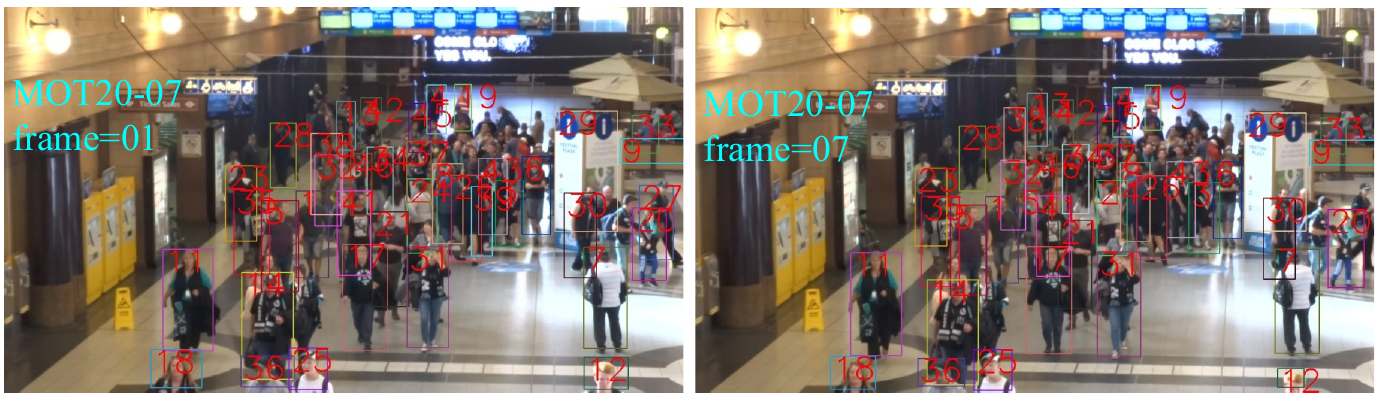}
  \caption{Visualizing the tracking performance of the OmniTrack method in dense crowds (MOT20~\cite{dendorfer2020mot20}).}
  \label{fig:MOT20}
\end{figure*}

\section{Discussion}
\subsection{Societal Impacts}
The OmniTrack framework is promising to enhance the safety and reliability of autonomous systems by improving Multi-Object Tracking (MOT) in panoramic settings, which is essential for applications such as self-driving cars and robots.
Its ability to process panoramic fields of view while mitigating distortions ensures robust performance in dynamic, real-world environments. 
These advancements have the potential to benefit a wide range of industries, particularly in navigation for individuals with visual impairments, drone-assisted rescue, and hazardous object detection. Furthermore, the development of the QuadTrack dataset, designed for high-speed sensor motion and panoramic field-of-view applications, fills a critical gap in available resources. 
Aim to make both the dataset and the associated code publicly available, we intend to accelerate progress in the field of omnidirectional multi-object tracking,
ultimately advancing the safety, efficiency, and inclusivity of automated systems in everyday life. Yet, it is inevitable that the deep model exhibits some false positives and negatives, and its practical deployment must account for the inherent uncertainty of deep neural networks. 
Additionally, while the technology is intended for benign applications, there exists a small risk of misuse, including potential military applications, and it may not be suitable for privacy-sensitive environments.

\subsection{Limitations and Future Work}
Although OmniTrack shows strong potential in the field of panoramic image tracking, it still has some limitations. 
While it does not exhibit ID confusion when targets are severely occluded, track loss can still occur in such scenarios. 
Future work could focus on addressing target occlusion, with one promising solution being multi-sensor fusion, such as integrating point cloud depth information to mitigate occlusion. 
This approach could extend 2D tracking to 3D tracking. Additionally, employing multiple agents that collaborate and share sensor information may enhance tracking performance,  ultimately reducing track loss caused by occlusion and improving overall system robustness.

\section{Visualization}

\noindent\textbf{MOT20.} OmniTrack is a MOT framework specifically designed for panoramic FoV, facilitating target localization and association across distorted and panoramic FoV images. Unlike pinhole cameras, where objects tend to be denser, panoramic images typically feature more sparsely distributed targets. To intuitively demonstrate OmniTrack's performance in dense pedestrian scenarios, we visualize its tracking results on sequence 07 of the MOT20 test set~\cite{dendorfer2020mot20}, as shown in Fig.~\ref{fig:MOT20}. The results indicate that OmniTrack successfully tracks most targets; however, it struggles with particularly small or heavily occluded objects, such as the one next to ID 20. The primary challenge stems from the limited training data in MOT20~\cite{dendorfer2020mot20}, which contains only $4$ sequences, posing a significant challenge for $OmniTrack_{Det}$. In future work, we aim to enhance tracking performance in dense target scenarios.

\begin{figure*}[!t]
  \centering
  \includegraphics[width=0.98\textwidth]{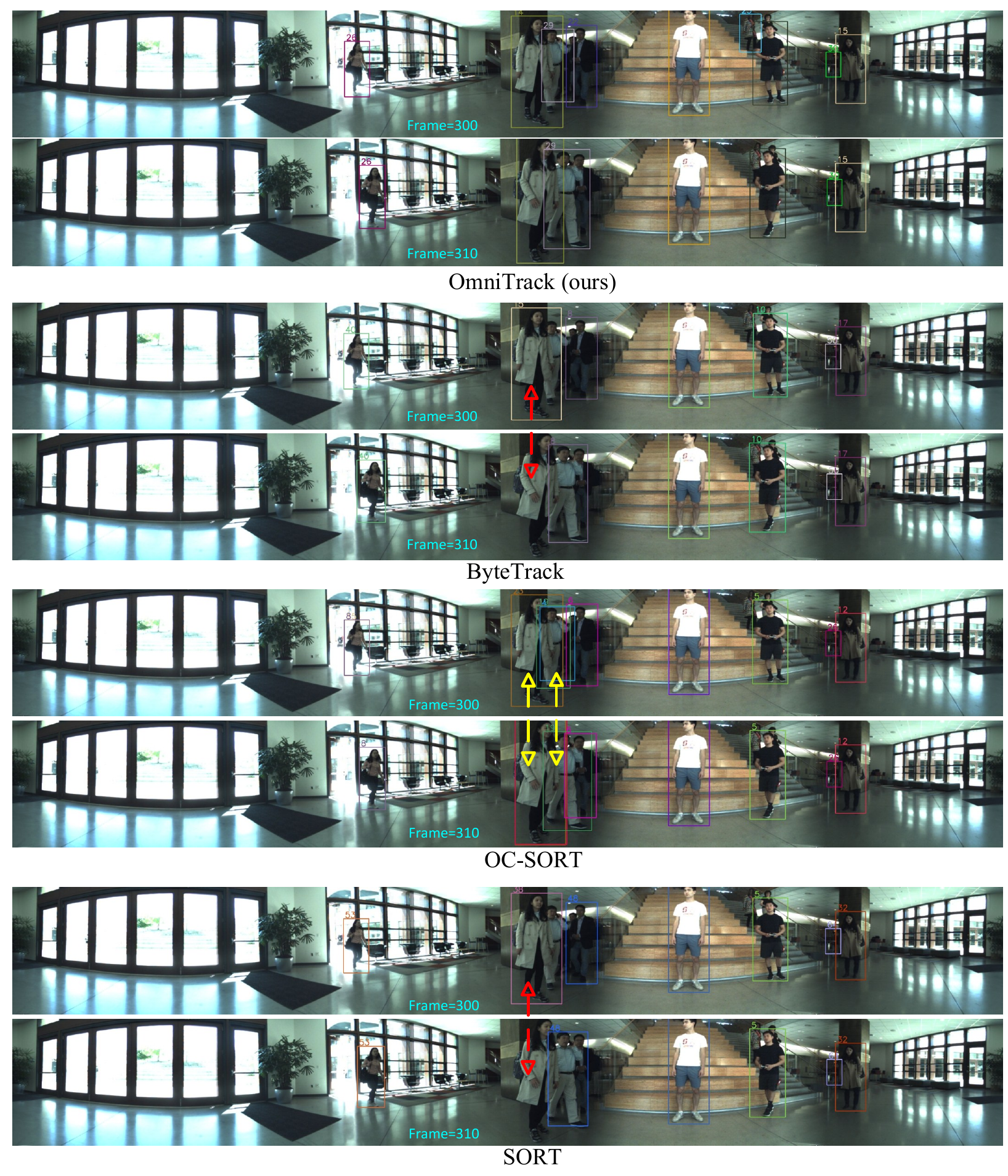}
  \caption{Visualization on the public JRDB dataset~\cite{martin2021jrdb}. The visualization compares the performance of OmniTrack, SOTA~\cite{bewley2016simple}, ByteTrack~\cite{zhang2022bytetrack}, and OC-SORT~\cite{cao2023observation} methods on the JRDB validation set. The \textcolor{red}{red} arrows in the figures indicate instances where the trajectories were not correctly tracked, leading to tracking losses, while \textcolor{yellow}{yellow} arrows highlight cases of track ID confusion, indicating ID switches.}
  \label{fig:vis_img_jrdb}
\end{figure*}

\begin{figure*}[!t]
  \centering
  \includegraphics[width=1\textwidth]{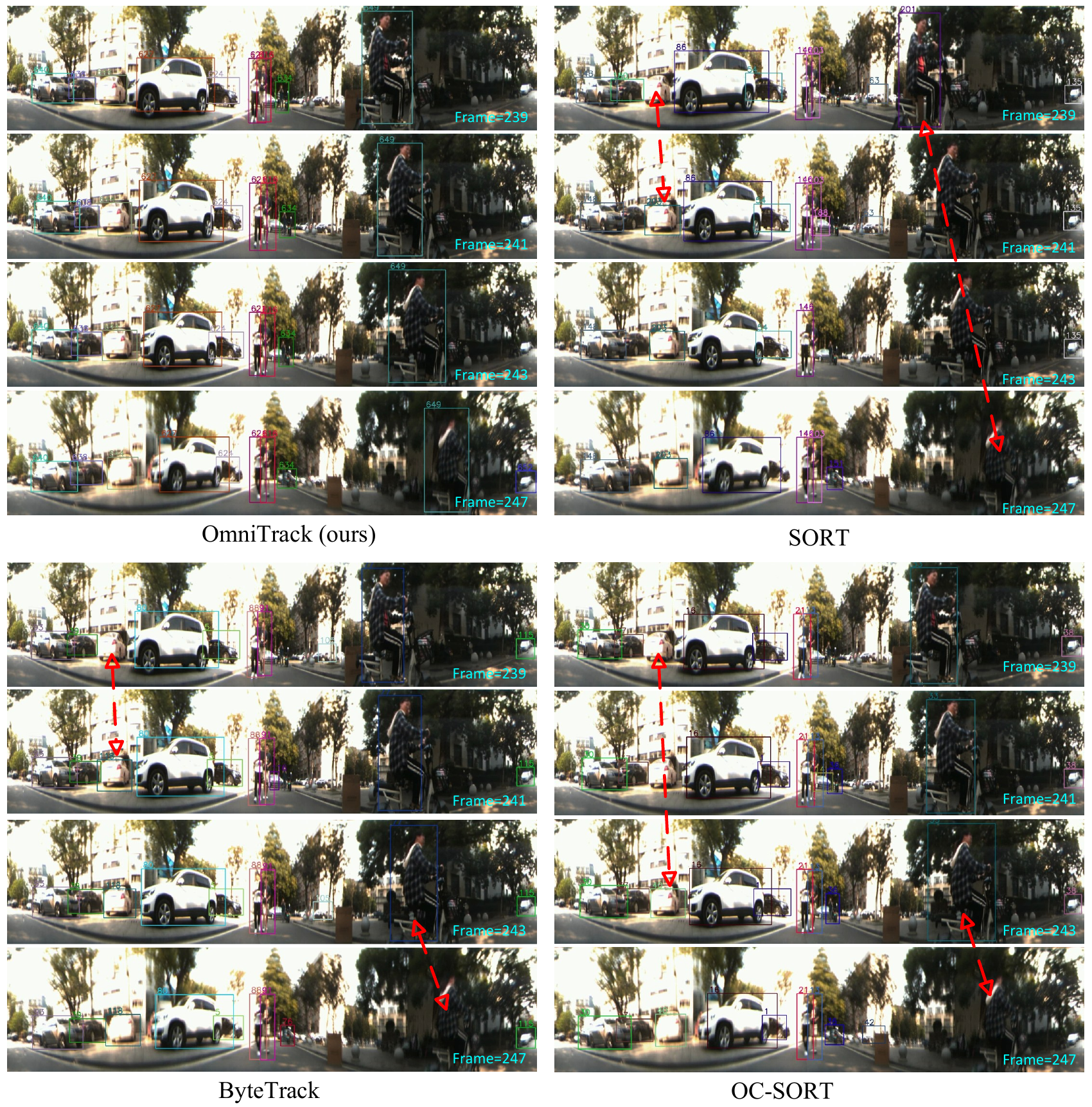}
  \caption{Visualization comparison on the established QuadTrack dataset. The visualization compares the performance of OmniTrack, SOTA~\cite{bewley2016simple}, ByteTrack~\cite{zhang2022bytetrack}, and OC-SORT~\cite{cao2023observation} methods on the QuadTrack test set. The \textcolor{red}{red} arrows in the figures indicate instances where the trajectories were not correctly tracked, leading to tracking losses.}
  \label{fig:vis_img_quadtrack}
\end{figure*}

\noindent\textbf{QuadTrack \& JRDB.} We visualize the final tracking results on the JRDB~\cite{martin2021jrdb} and QuadTrack datasets, as shown in Fig.~\ref{fig:vis_img_jrdb} and Fig.~\ref{fig:vis_img_quadtrack}. 
In these images, red arrows highlight instances where trajectories were lost and not correctly tracked, while yellow arrows indicate identity confusion, leading to ID switches. In Fig.~\ref{fig:vis_img_jrdb}, for the JRDB dataset~\cite{martin2021jrdb}, we observe that OmniTrack accurately tracks objects, even in scenes with a large number of people, without any ID switches or trajectory losses. 
In contrast, ByteTrack~\cite{zhang2022bytetrack} and SORT~\cite{bewley2016simple} both exhibit trajectory losses, while OC-SORT~\cite{cao2023observation} experiences multiple ID switches. 
In Figure~\ref{fig:vis_img_quadtrack}, for the QuadTrack dataset, the tracking of cyclists in the foreground remains intact, while OC-SORT, ByteTrack, and SORT all suffer from trajectory loss at frame $247$. These examples demonstrate OmniTrack's superior recall ability, further validating the effectiveness of our feedback mechanism and the FlexiTrack Instance in accurately maintaining targets in panoramic-FoV scenarios.

\end{document}